\documentclass[11pt]{article}
\PassOptionsToPackage{table}{xcolor}
\usepackage{custom}
\usepackage[T1]{fontenc}
\usepackage{microtype}
\usepackage{nicefrac}
\usepackage{url}
\usepackage{wrapfig}

\graphicspath{{Figures/}}

\newcommand{\best}[1]{\cellcolor{green!30}\textbf{#1}}
\newcommand{\second}[1]{\cellcolor{green!15}#1}
\newcommand{\DefinedAs}{\mathrel{\mathop:}=}

\title{Where to Spend Rollouts: Hit-Utility Optimal Rollout Allocation for Group-Based RLVR}
\author{
  Tao Wang$^{1}$ \quad
  Shuo Li$^{1}$\footnote{Now at Google DeepMind} \quad 
  Yan Sun$^{2}$ \quad
  Dongsheng Ding$^{3}$ \quad
  Edgar Dobriban$^{1}$ \\
  $^{1}$University of Pennsylvania \\
  $^{2}$New Jersey Institute of Technology\quad
  $^{3}$University of Tennessee, Knoxville
}

\begin{document}

\maketitle

\footnotetext[1]{\texttt{tawan@wharton.upenn.edu}, 
\texttt{lishuo1@seas.upenn.edu}, \texttt{dobriban@wharton.upenn.edu}}
\footnotetext[2]{\texttt{yan.sun@njit.edu}}
\footnotetext[3]
{\texttt{dongshed@utk.edu}}

\begin{abstract}
Reinforcement learning with verifiable rewards (RLVR) has emerged as a central paradigm for improving the reasoning capabilities of large language models. Group-based policy optimization methods, such as GRPO, typically allocate a fixed number of rollouts to every prompt. This uniform allocation can be inefficient: it over-allocates compute to prompts whose sampled groups are already saturated while under-exploring prompts for which additional samples may reveal useful correct trajectories. To address this limitation, we introduce \textit{hit utility}, the posterior probability that at least one rollout in a proposed additional allocation for a prompt will be correct. Building on this notion, we propose Hit-Utility Optimal Rollout Allocation (HORA), a learning-free rollout allocation policy that maximizes total posterior hit utility within each allocation batch. HORA adaptively reallocates rollout budgets while leaving the downstream reward evaluation and group-based advantage estimator unchanged. Across two training sets, four mathematical reasoning benchmarks, and three model families at multiple scales, HORA preserves comparable Pass@1 and improves Pass@$K$ over compute-matched GRPO, and it outperforms related allocation baselines at large $K$. It is also drop-in compatible with other group-based estimators such as RLOO. Ablation studies indicate that the uniform prior used by HORA is competitive with five prompt-conditioned learned-prior alternatives. 
\end{abstract}

\tableofcontents

\section{Introduction}

Reinforcement learning (RL) has become a central technique for improving the reasoning capabilities of large language models (LLMs)~\citep{guo2025deepseek}. Recent works have demonstrated that RL-based post-training can substantially enhance problem-solving performance, especially in domains where rewards can be verified automatically, such as mathematical reasoning and coding~\citep{wen2025reinforcement}. A widely used framework for reinforcement learning with verifiable rewards (RLVR) is group-based policy optimization, such as Group Relative Policy Optimization (GRPO)~\citep{shao2024deepseekmath}, which samples a group of rollouts per prompt and constructs a relative advantage estimate. This relative-advantage formulation avoids the need to train a high-quality value function, which can be computationally expensive \citep{shao2024deepseekmath, ahmadian2024back}. 

Although RL fine-tuning often improves overall accuracy, it has also been observed to degrade performance under Pass@$K$ evaluation \citep{yue2025does}, where success is defined as the event that at least one of $K$ independently sampled solutions is correct \citep{chen2021evaluating}. Pass@$K$ matters at both test time and training time: it is an upper bound on the accuracy of test-time methods that draw $K$ candidates and select one, such as best-of-$K$ selection with a verifier~\citep{snell2024scaling,brown2024monkeys}, and it reflects which problems remain learnable in further RLVR, since a problem on which no sampled solution is correct yields no learning signal. Since Pass@$K$ for large $K$ is dominated by the model's ability to solve challenging problems, such degradation under RL suggests a potential loss of capability on hard prompts. In group-based policy optimization methods, the same number of rollouts is typically generated for every prompt. This uniform rollout-allocation strategy can limit exploration of difficult prompts, where additional samples may be necessary to obtain even one correct solution, while wasting computation on prompts whose sampled groups are already nearly saturated \citep{nguyen2026adaptive}.

\begin{figure}[t]
\centering
\includegraphics[width=\linewidth]{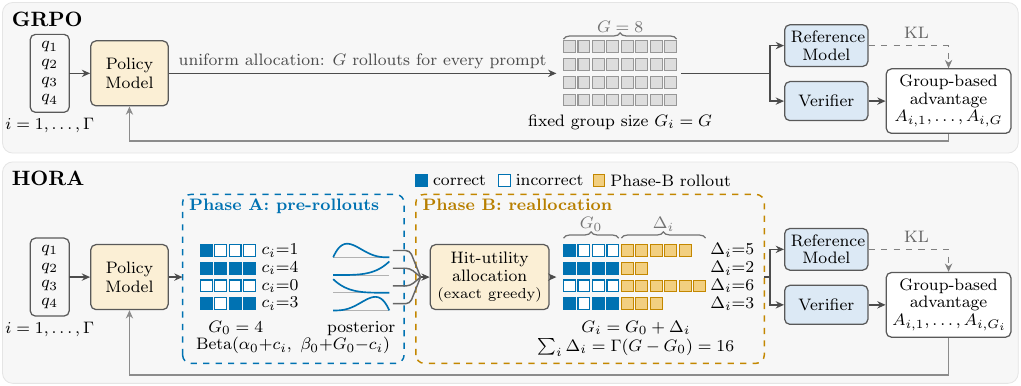}
\caption{Comparison of fixed-group GRPO (top) and HORA (bottom) on a batch of $\Gamma$ prompts. GRPO allocates a uniform group of size $G$ to every prompt. HORA instead splits the rollout layer into two stages: Phase~A draws $G_0$ pre-rollouts per prompt and forms a same-step posterior from the observed correct count $c_i$; Phase~B reallocates the remaining rollout budget $\Gamma(G-G_0)$ across the batch by maximizing the total hit utility, yielding $\Delta_i$ additional rollouts for prompt $q_i$ and a variable final group size $G_i = G_0 + \Delta_i$. The downstream loss update is then computed on the resulting variable-size groups (illustrated with $\Gamma=4$, $G_0=4$, $G=8$).}
\label{fig:pipeline}
\end{figure}

Existing work on mitigating Pass@$K$ degradation mainly operates at the \emph{reward- and advantage-shaping} layer. These methods modify how the training signal is computed or weighted, for example by adding an entropy term to the advantage estimate \citep{cheng2026reasoning} or by computing gradients only from high-entropy tokens \citep{wang2025beyond}. However, such approaches involve a trade-off between single-sample correctness and multi-sample coverage \citep{gai2025differential, wu2025invisible},
and methods based on entropy regularization are sensitive to the regularization strength~\citep{zhang2025revisiting}.

These limitations motivate us to look beyond reward and advantage shaping. In particular, we consider a different direction:
addressing Pass@$K$ degradation at the \emph{rollout-allocation} layer. Prior work on rollout allocation has mainly focused on training stability or computational efficiency \citep{qu2026can,qu2026small,nguyen2026adaptive}. In contrast, we show that appropriate rollout allocation can directly improve Pass@$K$ performance. Our proposed rollout allocation policy is architecturally orthogonal to reward- and advantage-shaping methods and can be combined with other group-based policy optimization frameworks.

We propose a learning-free rollout allocation policy, termed \underline{H}it-Utility \underline{O}ptimal \underline{R}ollout \underline{A}llocation (HORA), which prioritizes prompts whose posterior marginal hit utility remains large after accounting for diminishing returns, thereby shifting rollout budget toward the exploration frontier under a compute budget. To formalize this idea, we introduce the concept of \emph{hit utility}: the posterior probability that at least one additional rollout for a given prompt will be correct. At each training step, HORA performs allocation in two phases. First, it conducts a small number of pre-rollouts per prompt to estimate prompt-level success probabilities. Second, it distributes the remaining rollout budget across prompts so as to maximize the aggregate hit utility. Please refer to Figure~\ref{fig:pipeline} for a schematic overview of HORA and its comparison with GRPO. Our contributions are three-fold:

\begin{itemize}
\item We propose HORA, a novel rollout-allocation module that is compatible with group-based policy optimization methods. HORA uses the hit utility to prioritize prompts on the exploration frontier, and we show that the allocation that maximizes the total hit utility can be computed exactly by a greedy algorithm.

\item We conduct experiments across three model families at multiple scales, two training sets, and four benchmarks. HORA improves Pass@$K$ across a wide range of $K$, by up to $13.3$ points in AIME~2025 Pass@$1024$, while maintaining comparable Pass@1 to group-based policy optimization methods with default uniform allocation. These results demonstrate that HORA improves the trade-off between correctness and coverage.

\item We compare HORA with related allocation baselines and ablate its key design choices. HORA improves upon GRPO and outperforms these baselines at large $K$, and its uniform prior performs comparably to more sophisticated learned priors, demonstrating the simplicity and robustness of HORA.

\end{itemize}

\section{Related Work}
\label{sec:related}

\paragraph{Reinforcement Learning with Verifiable Rewards (RLVR).} In domains such as mathematical reasoning and code generation, rewards can be obtained automatically from verifiers, making RLVR a scalable post-training paradigm~\citep{guo2025deepseek,zeng2025simplerl,yang2025qwen25}. Group-based policy optimization methods, such as GRPO~\citep{shao2024deepseekmath}, Dr.~GRPO~\citep{liu2025understanding}, DAPO~\citep{yu2025dapo}, and RLOO~\citep{ahmadian2024back}, avoid learning a value function as in PPO~\citep{schulman2017proximal} and are widely used in this setting. Our work improves these methods by designing a more effective rollout-allocation policy.

\paragraph{Mitigating Pass@K Degradation in RLVR.}
Although RL fine-tuning improves overall accuracy, it has been observed to degrade performance under Pass@$K$ evaluation~\citep{yue2025does}. Existing work largely attributes this phenomenon to entropy collapse~\citep{he2025rewarding, yue2025does}. Proposed remedies control entropy, e.g., by adding an entropy term to the advantage~\citep{cheng2026reasoning}; restrict gradients to selected tokens or trajectories, e.g., updating only high-entropy tokens~\citep{wang2025beyond} or smoothing rewards only on correct trajectories~\citep{gai2025differential}; or reweight samples, e.g., upweighting negative samples~\citep{zhu2025surprising}. While these methods can help mitigate entropy collapse, they often involve a trade-off between correctness and coverage~\citep{gai2025differential}. Other methods reweight prompts in the loss by a Pass@$K$ or maximum-likelihood objective~\citep{chen2025passk,tajwar2026maximum} or replace the group baseline with a persistent value tracker~\citep{xu2025single}. All of these change how the signal from a given set of rollouts is weighted. In contrast, we focus on designing the rollout-allocation policy.

\paragraph{Adaptive Rollout Allocation.}
Existing dynamic-allocation methods for group-based policy optimization mainly focus on training efficiency and stability. Examples include filtering all-correct or all-incorrect groups~\citep{yu2025dapo}, predicting prompt difficulty with auxiliary models~\citep{qu2026can,qu2026small}, and framing rollout allocation as a variance-reduction problem for stable training~\citep{nguyen2026adaptive,fang2026allocate}.
Our work complements this direction by using rollout allocation as a mechanism for improving multi-sample reasoning performance, not only throughput, filtering, or variance reduction. Closest to our setting, DARS~\citep{yang2025depth} also targets Pass@$K$: it probes each prompt with a few rollouts and then adds rollouts to hard prompts. HORA differs from DARS in the allocation criterion, the solution, and the budget model: DARS raises the cumulative advantage of low-accuracy prompts with a difficulty-based schedule, whereas HORA maximizes a posterior hit utility that its greedy allocator solves exactly; and DARS tops up each prompt under a per-prompt cap, so its total rollout count varies, whereas HORA reallocates a fixed budget. An extended discussion of related work is given in Appendix~\ref{app:related}.

\section{HORA: Hit-Utility Optimal Rollout Allocation}
\label{sec:method}

\subsection{Background on RLVR}

We consider RLVR methods based on group-based advantage estimation, where advantages are computed relative to multiple rollouts sampled for the same prompt.
Let $\mathcal{Q}$ denote the prompt dataset and let $\pi_{\theta_{\mathrm{old}}}$ be the  rollout-generating policy. For a prompt $q \sim \mathcal{Q}$, $\pi_{\theta_{\mathrm{old}}}$ samples a group of $G$ responses $\{o_i\}_{i=1}^G$. 
The general
objective function for this class of methods is
\begin{equation}
\mathcal{J}(\theta)
=
\mathbb{E}_{q \sim \mathcal{Q},\, o_i\sim \pi_{\theta_{\mathrm{old}}}(\cdot \mid q)}
\Bigg[
\frac{1}{G}\sum_{i=1}^G
\frac{1}{|o_i|}\sum_{t=1}^{|o_i|}
\min\Big(
\rho_{i,t}(\theta)\hat{A}_{i,t},
\mathrm{clip}\big(\rho_{i,t}(\theta),1-\varepsilon,1+\varepsilon\big)\hat{A}_{i,t}
\Big)
\Bigg],
\label{eq:grpo-objective}
\end{equation}
where $\rho_{i,t}(\theta)
\DefinedAs
\frac{\pi_\theta(o_{i,t}\mid q,o_{i,<t})}
{\pi_{\theta_{\mathrm{old}}}(o_{i,t}\mid q,o_{i,<t})}$
is the per-token importance ratio, and $\hat{A}_{i,t}$ is the corresponding per-token advantage. In standard GRPO, after evaluating the reward vector $\mathbf{R}=(R(q,o_1),\ldots,R(q,o_G))$, the group-relative advantage is calculated as
$\hat{A}^{\text{GRPO}}_{i,t}
\; = \;
\frac{R(q,o_i)-\mathrm{mean}(\mathbf{R})}{\mathrm{std}(\mathbf{R})}$.
Here the mean and standard deviation are computed over the $G$ rewards of the responses; when the empirical standard deviation is zero, the group-relative advantage is set to zero, yielding no policy-gradient signal from that group.

While methods in this class differ in their specific advantage estimators and loss formulations (e.g., Dr.~GRPO~\citep{liu2025understanding}, which we use in our experiments, replaces $1/|o_i|$ in~\eqref{eq:grpo-objective} with a constant), they share a default rollout interface: each prompt receives a fixed group size $G$. This fixed-group design is limiting when prompt difficulty varies widely. Mathematically, if the probability of generating a correct rollout for a prompt is $p$, then the event that a group of size $G$ contains at least one correct rollout has probability $1-(1-p)^G$. This probability saturates quickly for easy prompts, while it remains small for prompts at the edge of the policy's exploration frontier.\footnote{More formally, for a test set with $n$ prompts, let $p_i$ denote the probability that the current policy generates a correct solution for problem $i$. Then the average $\mathrm{Pass@K}$ metric is $\mathrm{Pass@K} = \sum_{i=1}^{n} \left(1 - (1 - p_i)^K\right) / n$, therefore $1 - \mathrm{Pass@K} =  \sum_{i=1}^{n} (1 - p_i)^K / n$. Note that  $\sqrt[K]{ \sum_{i=1}^{n}(1-p_i)^K /n} \rightarrow \max_{i} 1- p_i$ as $K \rightarrow +\infty$. Therefore, for large $K$, $\mathrm{Pass@K}$ is dominated by problems with the smallest $p_i$, i.e., the hardest questions in the set.} A uniform group size therefore fails to account for heterogeneous sampling utility across prompts. 

\subsection{Posterior-guided rollout allocation}
\label{sec:HORA}

To address this bottleneck, we propose \underline{H}it-Utility \underline{O}ptimal \underline{R}ollout \underline{A}llocation (HORA), a learning-free rollout-allocation policy that modifies the rollout-allocation layer while keeping the downstream reward evaluation and group-based advantage estimation unchanged.
For a training batch of $\Gamma$ prompts $\{q_i\}_{i=1}^\Gamma$, standard training uses a fixed group size $G$, 
resulting in a total of $\Gamma G$ rollouts. 
HORA preserves the same total rollout budget but distributes it non-uniformly across prompts. Specifically, HORA divides the sampling process into two phases. Phase~A draws a uniform pre-rollout group of size $G_0 < G$ for every prompt. 
Phase~B re-allocates the remaining $\Gamma (G-G_0)$ rollouts across prompts according to the outcome of pre-rollouts. The resulting per-prompt group size satisfies $G_i=G_0+\Delta_i$ and $\sum_{i=1}^{\Gamma} \Delta_i=\Gamma\left(G-G_0\right)$,
where $\Delta_i \ge \,0$ is the number of additional rollouts to be allocated to prompt $q_i$ in Phase~B. HORA uses the Phase~A correctness counts to estimate each prompt's current success probability, and then determines the Phase~B allocation by maximizing total hit utility. Concretely, Phase~A and Phase~B proceed as follows.

\paragraph{Phase~A: Beta-Binomial model of prompt success rates.}
Let $R_{\mathrm{acc}}(q, o) \in \{0,1\}$ denote whether rollout $o$ yields a correct final answer for prompt $q$. For each prompt $q_i$, let $C_i$ denote the random Phase~A correctness count and let $c_i$ denote its observed realization. After drawing $G_0$ pre-rollouts $\{o_{i,j}^{\mathrm{pre}}\}_{j\,=\,1}^{G_0}$ from $\pi_{\theta_{\mathrm{old}}}$, we observe
$c_i=\sum_{j=1}^{G_0} R_{\mathrm{acc}}\left(q_i, o_{i, j}^{\mathrm{pre}}\right) \in\left\{0,1, \ldots, G_0\right\}$.
We model the prompt-level success probability $p_i$ with a Bayesian binomial model.
With a Beta prior $P_i \sim \mathrm{Beta}(\alpha_0,\beta_0)$ and binomial likelihood $C_i \mid P_i = p_i \sim \mathrm{Binomial}(G_0,p_i)$,
the posterior given the current step's pre-rollouts follows a $\mathrm{Beta}$ distribution by conjugacy:
\begin{equation}
P_i \mid C_i = c_i
\;\sim\; 
\mathrm{Beta}(\alpha_i, \beta_i),
\label{eq:posterior}
\end{equation}
where $\alpha_i = \alpha_0 + c_i$ and $\beta_i = \beta_0 + G_0 - c_i$. HORA uses the uniform prior $\alpha_0=\beta_0=1$ unless otherwise specified, under which the posterior mean $(c_i+1)/(G_0+2)$ is Laplace's rule of succession; this introduces no learned prompt-difficulty model and keeps no per-prompt state across steps.

\paragraph{Phase~B: hit-utility allocation.}
Given the posterior in~\eqref{eq:posterior}, we define the per-prompt \emph{hit utility} as
\begin{equation}
U_i^{\mathrm{hit}}(\Delta_i) 
\; \DefinedAs \; 
\mathbb{E}_{P_i \mid C_i = c_i}
\left[\, 
1 \,-\, (1 \,-\, P_i)^{\Delta_i}
\,\right],
\label{eq:hit-utility}
\end{equation}
which is the posterior probability that at least one of the $\Delta_i$ additional rollouts for prompt $q_i$ is correct, i.e., the posterior mean of Pass@$\Delta_i$ for $q_i$. HORA then allocates the remaining rollout budget by maximizing the total hit utility across prompts subject to the budget constraint:
\begin{equation}
\begin{split}
\max_{\Delta_1,\ldots,\Delta_{\Gamma} \, \in \, \mathbb{Z}_{\ge 0}}
\quad&
\sum_{i \,=\, 1}^{\Gamma} U_i^{\mathrm{hit}}(\Delta_i)
\\
\text{s.t.}
\quad&
\sum_{i \,=\, 1}^{\Gamma} \Delta_i \;=\; \Gamma\,(\, G \,-\, G_0\,).
\end{split}
\label{eq:allocation-problem}
\end{equation}

\paragraph{Greedy allocator.}
The optimization problem \eqref{eq:allocation-problem} can be solved exactly by a greedy allocation rule. Let $M_i(\ell)$ denote the \emph{marginal hit utility} of assigning the $(\ell+1)$-th additional rollout to prompt $q_i$:
\begin{equation}
M_i(\ell) 
\;=\; 
U_i^{\mathrm{hit}}(\ell + 1) \,-\, U_i^{\mathrm{hit}}(\ell) 
\;=\; 
\frac{B(\alpha_i + 1, \beta_i + \ell)}{B(\alpha_i, \beta_i)},
\label{eq:marginal}
\end{equation}
where $B(\cdot,\cdot)$ denotes the beta function. In implementation, the same marginal can be updated stably using $M_i(0)=\alpha_i/(\alpha_i+\beta_i)$ and $M_i(\ell+1)=M_i(\ell)(\beta_i+\ell)/(\alpha_i+\beta_i+\ell+1)$, avoiding repeated beta-function evaluations. The greedy rule starts from $\Delta_i=0$ for all prompts and repeatedly assigns the next rollout to the prompt with the largest current marginal gain $M_i(\Delta_i)$ until the Phase~B budget $\Gamma (G-G_0)$ is exhausted. Algorithm~\ref{alg:hora} in Appendix~\ref{app:algorithm} summarizes the complete rollout construction procedure, and the following proposition establishes optimality of its Phase~B allocation.

\begin{proposition}\label{prop:greedy-optimal}
The Phase~B allocation returned by Algorithm~\ref{alg:hora} is a globally optimal solution to the integer allocation problem~\eqref{eq:allocation-problem}.
\end{proposition}

The proof is provided in Appendix~\ref{app:greedy-proof}. The key observation is that each marginal sequence $M_i(0),M_i(1),\ldots$ is decreasing, so the optimal allocation is obtained by selecting the largest $\Gamma (G-G_0)$ marginal gains across prompts.

\begin{remark}\label{rem:posterior}
Using the posterior distribution rather than the plug-in estimate $C_i/G_0$ is important for allocation. Under the plug-in estimate, any prompt with $c_i=0$ has zero estimated success probability and therefore zero marginal hit utility, causing it to receive no additional rollouts in Phase~B except in degenerate cases where all remaining marginal hit utilities are also zero. The uniform Beta prior avoids this zero-probability degeneracy while remaining learning-free. A formal statement is given in Appendix~\ref{app:plugin-degeneracy}. We also consider the prompt-conditioned learned priors in Section~\ref{sec:ablations}, but these variants do not provide consistent gains in our experiments.
\end{remark}

\paragraph{Training objective with variable-size groups.} After allocation, the additional rollouts are drawn from $\pi_{\theta_{\mathrm{old}}}$ and pooled with the pre-rollouts to form a final group of size $G_i=G_0+\Delta_i$ for prompt $q_i$. Reward evaluation and group-based advantage estimation are computed on this group. In the policy loss of~\eqref{eq:grpo-objective}, we replace $1/G$ with $1/G_i$ so that each prompt has equal weight regardless of how many additional rollouts it receives. Thus, HORA changes only the rollout-allocation layer and remains compatible with RLVR methods based on group-based advantage estimation.

\section{Experiments}
\label{sec:results}

We evaluate the benefits of HORA on mathematical reasoning across model families, model scales, training sets, and benchmark difficulty. Using standard practice~\citep{gai2025differential, zhu2025surprising}, we experiment with
three Qwen2.5 backbones up to 7B parameters: Qwen2.5-1.5B-Instruct, Qwen2.5-3B, and Qwen2.5-7B~\citep{yang2025qwen25}.
Each model is fine-tuned on MATH12k~\citep{hendrycks2021measuring} 
and evaluated on four benchmarks: MATH500~\citep{hendrycks2021measuring}, AMC23~\citep{cao2025step}, AIME~2024~\citep{patel2024aime}, and AIME~2025~\citep{ospanov2025hermes}. To assess generality across model families and training sets, we further train Llama-3.2-3B-Instruct and Qwen3-1.7B-Base on MATH12k, and Qwen3-1.7B-Base on the harder DAPO-17k~\citep{yu2025dapo}, and evaluate them on AIME~2025. During training, each rollout uses a verifiable reward $(R_{\mathrm{fmt}}(q, o)  + R_{\mathrm{acc}}(q, o)) / 2$,
where $R_{\mathrm{fmt}}, R_{\mathrm{acc}} \in \{0,1\}$. The format reward $R_{\mathrm{fmt}}$ checks whether the completion follows the required \texttt{<think>...</think><answer>...</answer>} structure, with reasoning enclosed in \texttt{<think>} tags and the final answer enclosed in \texttt{<answer>} tags. The accuracy reward $R_{\mathrm{acc}}$ checks whether the extracted answer is correct. All evaluation metrics are computed using $R_{\mathrm{acc}}$.

For each prompt $q$ in the evaluation dataset $\mathcal{D}$, we draw a pool of $N$ independent samples, $o_1,\ldots,o_N$, from the evaluated policy, where $r_j(q)=R_{\mathrm{acc}}(q,o_j)$ denotes the correctness of the $j$-th sample. In line with \cite{yue2025does}, we assess reasoning performance using the Pass@1 and Pass@$K$ metrics. Specifically, for any $K \le N$, 
we employ the low-variance unbiased estimator introduced by \citet{chen2021evaluating}:
\begin{equation}
    \widehat{\mathrm{Pass}}@K 
    \; \DefinedAs \;
    \mathbb{E}_{q \,\sim\, \mathcal{D}}\!\left[1 - \binom{N - c(q)}{K}\bigg/\binom{N}{K}\right],\label{eq:pass-k}
\end{equation}
where $c(q) \DefinedAs \sum_{j=1}^{N} r_j(q)$ represents the total number of correct samples within the evaluation pool for prompt $q$. 
To accommodate varying benchmark difficulties, we follow the standard practice~\citep{yue2025does} and tailor our sample sizes and 
evaluation range.
For MATH500 and AMC23, we set $N=256$ and evaluate Pass@$K$ for $K \in \{8,16,32,64,128,256\}$. For the AIME~2024 and AIME~2025 datasets, we scale the pool size to $N=1024$ and evaluate across $K \in \{8,16,32,64,128,256,512,1024\}$.

We use GRPO as a baseline
and compare with HORA on top of GRPO (denoted as GRPO + HORA in tables and figures). Our implementation uses the Dr.~GRPO length normalization; we refer to the compute-matched fixed-allocation baseline as GRPO for readability.
We set the group size to $G=32$ and the HORA pre-rollout count to $G_0=8$. We set the KL penalty coefficient to zero throughout. For a training batch of $\Gamma$ prompts $\{q_i\}_{i=1}^\Gamma$, 
HORA and  GRPO use the same total rollout budget per training step, $\Gamma G$, so the comparison is compute-matched and differs only in how rollouts are allocated across prompts. We compare HORA against  GRPO with the same group size and against the corresponding pre-RL checkpoint, denoted as ``Base''.\footnote{For the 1.5B experiments, we use \texttt{Qwen2.5-1.5B-Instruct} as the base checkpoint because \texttt{Qwen2.5-1.5B} rarely follows the required output format.} Per-model training configuration details are listed in Appendix~\ref{app:configs}. 

\subsection{Main Results}
\label{sec:main-results}

\begin{figure}[ht]
\centering
\includegraphics[width=0.9\linewidth]{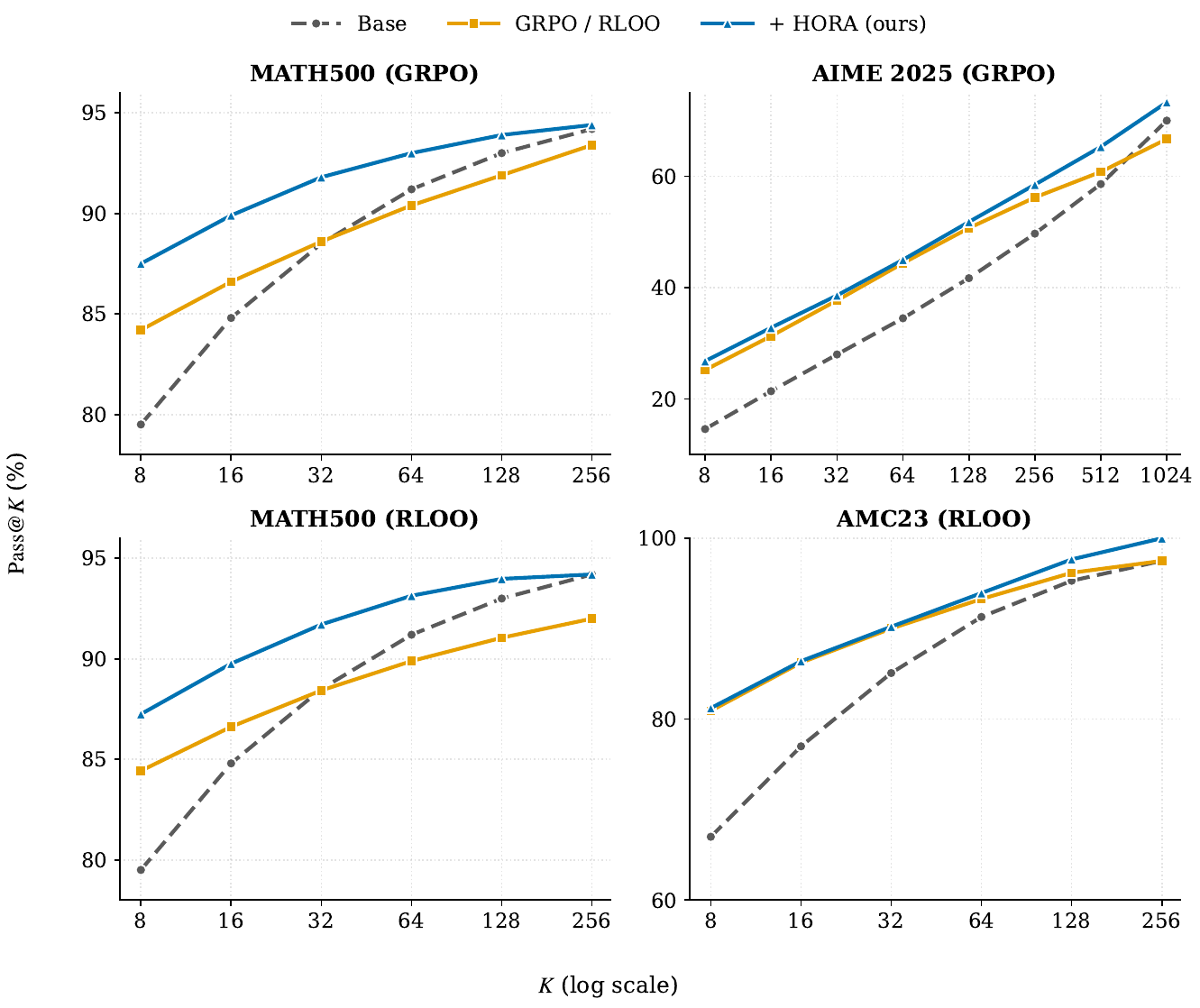}
\caption{Pass@$K$ curves on Qwen2.5-7B. Top: GRPO and GRPO + HORA on in-distribution MATH500 and out-of-distribution AIME~2025; HORA (blue) lies strictly above the Base checkpoint (gray) at every $K$ we evaluate, whereas GRPO (orange) falls below it on MATH500 from $K=64$ onward and on AIME~2025 at $K=1024$ ($66.7$ vs.\ $70.0$). Bottom: the same comparison with RLOO as the advantage estimator, on MATH500 and AMC23.}

\label{fig:passk-aime}
\end{figure}

Table~\ref{tab:main-results} reports
estimated
Pass@1 and Pass@$K$ across the twelve Qwen2.5 model-benchmark configurations. Figure~\ref{fig:passk-aime} (top) presents representative Pass@$K$ curves for Qwen2.5-7B on MATH500 and AIME~2025; the remaining Pass@$K$ curves are provided in Appendix~\ref{app:full-passk}. 

\begin{table}[ht]
\centering
\caption{Pass@1 and Pass@$K$ (\%) across Qwen2.5 model scales and four benchmarks. We use $K=256$ for MATH500 and AMC23, and $K=1024$ for AIME~2024 and AIME~2025. The best value per column within each model scale is highlighted in \textcolor{green!70!black}{green}.}
\label{tab:main-results}
\vspace{-0.5em}
\small
\setlength{\tabcolsep}{4pt}
\begin{tabular}{llcccccccc}
\toprule
\multirow{2}{*}{Scale} & \multirow{2}{*}{Method}
 & \multicolumn{2}{c}{MATH500} & \multicolumn{2}{c}{AMC23} & \multicolumn{2}{c}{AIME~2024} & \multicolumn{2}{c}{AIME~2025} \\
\cmidrule(lr){3-4} \cmidrule(lr){5-6} \cmidrule(lr){7-8} \cmidrule(lr){9-10}
 & & Pass@1 & Pass@$K$ & Pass@1 & Pass@$K$ & Pass@1 & Pass@$K$ & Pass@1 & Pass@$K$ \\
\midrule
\multirow{3}{*}{1.5B-Inst.}
 & Base    & 46.5          & \best{91.4}   & 22.0          & 92.5          & 2.2          & \best{66.7} & 0.9          & \best{63.3} \\
 & GRPO    & \best{50.8}   & 89.4          & \second{27.4} & 87.5          & \best{3.1}   & 53.3        & \best{1.0}   & 56.7 \\
 & GRPO + HORA & \second{50.6} & \second{90.4} & \best{27.8}   & \best{95.0}   & \second{3.0} & 63.3        & \best{1.0}   & 60.0 \\
\midrule
\multirow{3}{*}{3B}
 & Base    & 27.6          & \best{93.6}   & 11.1          & \best{100.0} & 1.2          & 53.3        & 0.7          & 56.7 \\
 & GRPO    & \best{59.9}   & 91.8          & \second{35.9} & 97.5         & \best{6.2}   & 56.7        & \best{2.4}   & 63.3 \\
 & GRPO + HORA & \second{59.1} & 92.0          & \best{36.3}   & 97.5         & \second{5.7} & \best{60.0} & \second{2.0} & \best{70.0} \\
\midrule
\multirow{3}{*}{7B}
 & Base    & 41.9          & \second{94.2} & 24.5          & 97.5         & 4.8           & 70.0        & 2.6          & 70.0 \\
 & GRPO    & 70.8          & 93.4          & 49.8          & \best{100.0} & \second{11.7} & 70.0        & \second{8.3} & 66.7 \\
 & GRPO + HORA & \best{73.9}   & \best{94.4}   & \best{53.9}   & 97.5         & \best{12.0}   & \best{73.3} & \best{9.2}   & \best{73.3} \\
\bottomrule
\end{tabular}
\end{table}

\vspace{-0.5em}

\paragraph{HORA achieves a superior trade-off between Pass@1 and Pass@$K$.}
In terms of Pass@1, 
\emph{HORA outperforms GRPO on all four benchmarks at the 7B} 
scale with gains of $+3.1$ points on MATH500 and $+4.1$ points on AMC23, while performing comparably at the 1.5B and 3B scales. These findings suggest that HORA preserves the average correctness achieved by standard GRPO and may even enhance it for sufficiently large models.

Furthermore, 
the Pass@$K$ metrics demonstrate a more consistent advantage over GRPO. 
\emph{Across the twelve model-benchmark configurations, HORA improves upon GRPO on ten 
settings} and matches its performance on one; the single observed decrease occurs on Qwen2.5-7B AMC23, where GRPO has already saturated at Pass@$256=100.0$.
The most substantial gains emerge on benchmarks where additional exploration is highly beneficial. For instance, HORA improves Pass@$K$ by $+10.0$ points on Qwen2.5-1.5B AIME~2024, $+7.5$ points on Qwen2.5-1.5B AMC23, $+6.7$ points on Qwen2.5-3B AIME~2025, and $+6.6$ points on Qwen2.5-7B AIME~2025.
On the four AIME configurations at the 3B and 7B scales, HORA also surpasses the corresponding Base model. 

As illustrated in Figure~\ref{fig:passk-aime} (top), this behavior persists across the full Pass@$K$ curve. While GRPO often drops below the Base model at large values of $K$, HORA consistently remains above it in representative 7B settings. At the 1.5B scale, the Base model remains strongest in Pass@$K$ across three of the four benchmarks, indicating limited exploration headroom for the smallest instruction-tuned models. Nevertheless, HORA  bridges this gap, performing closer to the Base model than GRPO across every 1.5B evaluation.

On other model families and training sets (Table~\ref{tab:families}), HORA likewise keeps Pass@1 comparable to GRPO, while AIME~2025 Pass@$1024$ improves by $6.7$ points for Llama-3.2-3B-Instruct and Qwen3-1.7B-Base trained on MATH12k and by $13.3$ points for Qwen3-1.7B-Base trained on DAPO-17k (full Pass@$K$ curves in Appendix~\ref{app:full-passk}).

\begin{table}[ht]
\centering
\caption{Pass@1 and Pass@$K$ (\%) on AIME~2025 for other model families and training sets, with $K=1024$. The best Pass@1 and Pass@$K$ in each setting are in bold.}
\label{tab:families}
\setlength{\tabcolsep}{4pt}
\begin{tabular}{@{}llcc@{}}
\toprule
Setting & Method & Pass@1 & Pass@$K$ \\
\midrule
Llama-3.2-3B-Inst. & Base & 0.2 & 53.3 \\
(MATH12k) & GRPO & \textbf{0.6} & 50.0 \\
 & GRPO + HORA & 0.5 & \textbf{56.7} \\
\midrule
Qwen3-1.7B-Base & Base & 1.9 & 56.7 \\
(MATH12k) & GRPO & \textbf{4.3} & 63.3 \\
 & GRPO + HORA & 3.9 & \textbf{70.0} \\
\cmidrule(l){2-4}
(DAPO-17k) & GRPO & \textbf{4.1} & 56.7 \\
 & GRPO + HORA & 3.9 & \textbf{70.0} \\
\bottomrule
\end{tabular}
\end{table}

\paragraph{HORA is compatible with GRPO-family objectives.}
As discussed, HORA should be drop-in compatible with other prompt-normalized GRPO-family advantage estimators.
To investigate this, we stack HORA on top of REINFORCE Leave-One-Out (RLOO)~\citep{ahmadian2024back} on Qwen2.5-7B and compare it with a compute-matched RLOO baseline using fixed group size $G=32$.
Figure~\ref{fig:passk-aime} (bottom) shows that HORA preserves the same qualitative effect under RLOO. 
On MATH500, RLOO improves Pass@$K$ over Base at small $K$ but falls below Base at large $K$, while RLOO + HORA remains above both curves across the evaluated range. On AMC23, Base catches up with the RLOO baseline at high $K$, whereas RLOO + HORA remains higher and reaches Pass@$256=100.0$. These results support that HORA can transfer to other advantage estimators.

\begin{figure}[ht]
\centering
\includegraphics[width=0.9\linewidth]{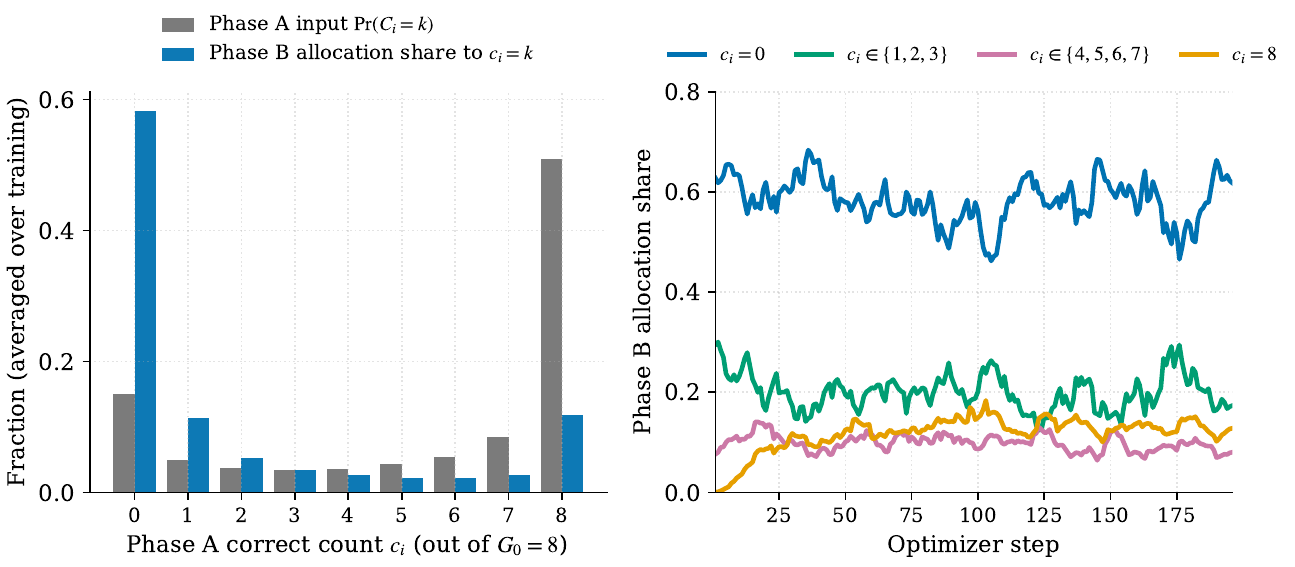}
\caption{Allocation dynamics on the Qwen2.5-7B run. Left: average Phase~A input fraction $\Pr(C_i=k)$ and average Phase~B allocation share for each observed pre-rollout correct-count bucket $c_i=k$, averaged over $196$ optimizer steps. Right: Phase~B allocation share over training, grouped by $c_i=0$, $c_i \in \{1,2,3\}$, $c_i \in \{4,5,6,7\}$, and $c_i=8$; curves are smoothed with a $6$-step moving average.}
\label{fig:allocation-analysis}
\end{figure}

\paragraph{HORA effectively reallocates rollout budgets.}
We inspect how HORA allocates the Phase~B budget. Figure~\ref{fig:allocation-analysis} shows that the pre-rollout outcomes are highly non-uniform: many prompts are already saturated after Phase~A, while a smaller fraction receive no correct pre-rollouts. HORA reallocates budget toward hard prompts. Averaged over training, prompts with $c_i=0$ account for $15\%$ of the Phase~A input distribution but receive $58\%$ of the Phase~B budget. Conversely, prompts with $c_i = G_0 = 8$ account for $51\%$ of the input distribution but receive only $12\%$ of the Phase~B budget. This nonzero allocation to $c_i=8$ is expected under the marginal utility rule: an all-correct pre-rollout group has a high first marginal hit probability, but its subsequent marginal utilities drop quickly. These additional rollouts also reveal residual incorrect responses in otherwise all-correct pre-rollout groups and thereby restore nonzero group-relative signal. Likewise, for prompts with $c_i=0$, the extra rollouts can supply the correct completions that an all-incorrect pre-rollout group lacks: on average, such a prompt receives $104$ extra rollouts and ends with $3.6$ correct ones. Over training,
the allocation to $c_i=8$ grows gradually as the policy improves and more prompts become saturated after Phase~A. HORA also produces longer responses on average and maintains a heavier entropy tail than GRPO, suggesting that the allocation policy preserves more exploratory sampling behavior; details
are reported in Appendix~\ref{app:length-stats}.

\vspace{-0.5em}

\begin{wrapfigure}[18]{r}{0.48\linewidth}
\centering
\includegraphics[width=\linewidth]{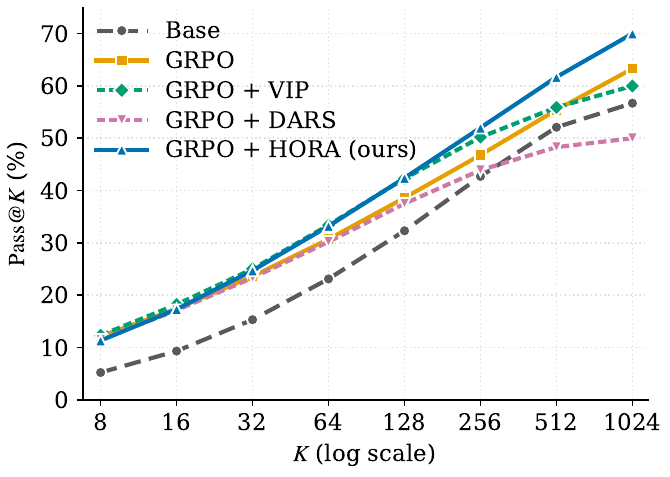}
\caption{AIME~2025 Pass@$K$ of Qwen2.5-3B trained on MATH12k under different rollout-allocation rules: uniform allocation (GRPO), the variance-minimizing rule of VIP, the difficulty-adaptive sampling rule of DARS, and HORA.}
\label{fig:baselines}
\end{wrapfigure}
\paragraph{HORA outperforms related allocation baselines in Pass@$K$.}
We compare HORA with two alternative rules for spending the Phase~B budget, both run inside the same allocation layer when training Qwen2.5-3B on MATH12k: the variance-minimizing rule of VIP~\citep{nguyen2026adaptive} and the difficulty-adaptive sampling rule of DARS~\citep{yang2025depth}, with its published schedule ($G_0=8$, per-prompt cap $32$). On AIME~2025 (Figure~\ref{fig:baselines}), all rules are close at small $K$, and VIP and HORA remain comparable up to $K=128$; from $K=256$ on, HORA leads and the gap grows with $K$, reaching $70.0$ at $K=1024$, against $63.3$ for GRPO and $60.0$ for VIP. DARS ends lowest ($50.0$); its schedule does not guarantee that the full rollout budget is spent, which may contribute to its lower Pass@$K$.

\paragraph{HORA adds little wall-clock overhead.}
The extra synchronization between the two phases adds little wall-clock time: on average, an optimizer step takes $427$\,s for GRPO and $457$\,s for HORA on Qwen2.5-3B ($+6.9\%$), and $591$\,s and $582$\,s on Qwen2.5-7B; the allocator itself takes $21$\,ms per step. A stage-wise breakdown is given in Appendix~\ref{app:cost}.

\subsection{Ablations}
\label{sec:ablations}
We next ablate the main design choices in HORA. These experiments address three questions: 
(1) is hit-utility allocation
needed beyond simpler hard-first heuristics,
(2) how much budget should be spent on pre-rollouts and how large the total group size $G$ should be, and
(3) do auxiliary prompt-difficulty predictors improve over the fixed-prior posterior used by HORA.
Unless otherwise stated, all ablations are conducted on Qwen2.5-3B with the same total rollout budget as the main experiments.

\paragraph{Comparison with the hard-first heuristic.}
We ablate two design choices in HORA on Qwen2.5-3B (Table~\ref{tab:allocator-g0}). The first ablation tests a natural hard-first heuristic motivated by the limitation of uniform group size: after the $G_0$ pre-rollouts, allocate the remaining budget only to prompts with $c_i=0$, i.e., prompts for which all pre-rollouts are incorrect. This heuristic performs worse than the hit-utility allocator, reducing Pass@1 on all four benchmarks and lowering AIME~2025 Pass@$K$ from $70.0$ to $56.7$. 
This supports that marginal hit utility brings benefits beyond simply routing the budget to the hardest prompts.

\begin{table}[ht]
\centering
\caption{Hard-first and pre-rollout-budget ablations on Qwen2.5-3B. The hard-first heuristic allocates the Phase~B budget only to prompts with $c_i=0$ after the $G_0$ pre-rollouts. The $G_0=16$ variant doubles the pre-rollout budget while keeping the total group size fixed at $G=32$. Pass@1 and Pass@$K$ (\%) follow the evaluation setup in Section~\ref{sec:results}.}
\label{tab:allocator-g0}
\vspace{3pt}
\setlength{\tabcolsep}{4pt}
\begin{tabular}{@{}lcccccccc@{}}
\toprule
\multirow{2}{*}{Variant}
 & \multicolumn{2}{c}{MATH500} & \multicolumn{2}{c}{AMC23} & \multicolumn{2}{c}{AIME~2024} & \multicolumn{2}{c}{AIME~2025} \\
\cmidrule(lr){2-3} \cmidrule(lr){4-5} \cmidrule(lr){6-7} \cmidrule(lr){8-9}
 & Pass@1 & Pass@$K$ & Pass@1 & Pass@$K$ & Pass@1 & Pass@$K$ & Pass@1 & Pass@$K$ \\
\midrule
Hard-first & 56.1          & 91.2          & 32.7          & \textbf{97.5} & 5.6          & \textbf{60.0} & 1.7          & 56.7          \\
HORA ($G_0{=}8$)    & 59.1          & \textbf{92.0} & \textbf{36.3} & \textbf{97.5} & \textbf{5.7} & \textbf{60.0} & \textbf{2.0} & 70.0          \\
HORA ($G_0{=}16$)   & \textbf{59.5} & 91.2          & 36.0          & 92.5          & 5.5          & \textbf{60.0} & 1.8          & \textbf{73.3} \\
\bottomrule
\end{tabular}
\end{table}

\paragraph{Pre-rollout budget $G_0$ and total group size $G$.} The second ablation doubles the pre-rollout budget from $G_0=8$ to $G_0=16$ while keeping the total group size fixed at $G=32$. Increasing $G_0$ provides more within-step evidence about prompt difficulty, but it also reduces the adaptive Phase~B budget from $24$ to $16$ rollouts per prompt on average. Empirically, $G_0=16$ gives similar Pass@1 but does not uniformly improve Pass@$K$: it improves AIME~2025 ($70.0 \to 73.3$) while reducing AMC23 ($97.5 \to 92.5$). We therefore use $G_0=8$ as the default trade-off between estimating prompt difficulty and retaining budget for adaptive allocation. We also compare GRPO and HORA at total group sizes $G=32$ and $G=16$, with $G_0=8$ in both settings. Reducing $G$ to $16$ cuts the adaptive Phase~B budget from $24$ to $8$ rollouts per prompt on average. At $G=16$, HORA and GRPO are close at every $K$ on AIME~2025 (Pass@1 $2.1$ vs.\ $1.7$; Pass@$1024$ $60.0$ for both), so HORA needs a reasonably large group size compared to the pre-rollout budget $G_0$.

\paragraph{Prior design.}
We next test whether the fixed $\mathrm{Beta}(1,1)$ prior in HORA can be improved by incorporating prompt-conditioned prior estimates.
We consider both parametric and nonparametric alternatives:
(a) linear probes over frozen base-model hidden states and
(b) linear probes over frozen MPNet embeddings \citep{song2020mpnet},
as well as
(c)
the Gaussian-process predictor from VIP~\citep{nguyen2026adaptive};
see Appendix~\ref{app:prior-details} for details.
These alternatives are more expressive, but they also introduce additional complexity. Parametric probes must be fitted or updated online from noisy rollout outcomes, making their quality sensitive to the choice of update schedule, learning rate, and distribution shift during RL training. Nonparametric predictors such as the Gaussian process used in VIP~\citep{nguyen2026adaptive} require maintaining a kernel matrix over training-prompt embeddings and performing recursive covariance updates, introducing memory and computation costs that scale with the prompt set size.

Figure~\ref{fig:prior-full} in Appendix~\ref{app:prior-details} shows that these more complex priors do not yield a clear advantage over the fixed prior. On AMC23, several prompt-conditioned priors are slightly ahead at intermediate values of $K$, but the fixed prior catches up by $K=256$. On AIME~2025, the fixed prior is consistently better across the evaluated range and reaches Pass@$1024=70.0$, compared with at most $66.7$ among the predictor-based variants. These results suggest that, in our current setting, the same-step pre-rollout outcomes already provide a strong allocation signal.
Designing prompt-conditioned priors that reliably improve over this simple baseline remains an open question.

\section{Conclusion and Discussion}

We introduce HORA, a novel rollout-allocation policy that prioritizes prompts near the exploration frontier in group-based RLVR training.
Across model families, scales, training sets, and mathematical reasoning benchmarks, HORA
improves the trade-off between correctness, measured by Pass@1, and multi-sample coverage, measured by Pass@$K$, compared with GRPO and with related allocation baselines.
These results suggest that, even without modifying the reward model,
advantage estimator, or policy architecture, adaptive rollout allocation can serve as a layer for mitigating Pass@$K$ degradation under fixed compute budgets. The rollout-allocation view also provides a flexible module for future extensions, including larger-scale training, refined prior estimation, and alternative utility specifications.

\section*{Acknowledgments}

This work was partially supported by the NSF, ARO, AFOSR, ONR, and the Sloan Foundation. Portions of this work was performed on High Performance Computing resources provided by the Advanced Research Computing Services team at NJIT.

\newpage

\bibliographystyle{plainnat}
\bibliography{ref}

\begin{thebibliography}{33}
\providecommand{\natexlab}[1]{#1}
\providecommand{\url}[1]{\texttt{#1}}
\expandafter\ifx\csname urlstyle\endcsname\relax
  \providecommand{\doi}[1]{doi: #1}\else
  \providecommand{\doi}{doi: \begingroup \urlstyle{rm}\Url}\fi

\bibitem[Ahmadian et~al.(2024)Ahmadian, Cremer, Gall{\'e}, Fadaee, Kreutzer,
  Pietquin, {\"U}st{\"u}n, and Hooker]{ahmadian2024back}
Arash Ahmadian, Chris Cremer, Matthias Gall{\'e}, Marzieh Fadaee, Julia
  Kreutzer, Olivier Pietquin, Ahmet {\"U}st{\"u}n, and Sara Hooker.
\newblock Back to basics: Revisiting reinforce-style optimization for learning
  from human feedback in {LLMs}.
\newblock In \emph{Proceedings of the 62nd Annual Meeting of the Association
  for Computational Linguistics (Volume 1: Long Papers)}, pages 12248--12267,
  2024.

\bibitem[Brown et~al.(2024)Brown, Juravsky, Ehrlich, Clark, Le, R{\'e}, and
  Mirhoseini]{brown2024monkeys}
Bradley Brown, Jordan Juravsky, Ryan Ehrlich, Ronald Clark, Quoc~V. Le,
  Christopher R{\'e}, and Azalia Mirhoseini.
\newblock Large language monkeys: Scaling inference compute with repeated
  sampling.
\newblock \emph{arXiv preprint arXiv:2407.21787}, 2024.

\bibitem[Cao et~al.(2025)Cao, Zou, Peng, Chen, Ning, and Li]{cao2025step}
Lang Cao, Yingtian Zou, Chao Peng, Renhong Chen, Wu~Ning, and Yitong Li.
\newblock Step guided reasoning: Improving mathematical reasoning using
  guidance generation and step reasoning.
\newblock In \emph{Proceedings of the 2025 Conference on Empirical Methods in
  Natural Language Processing}, pages 21112--21129, 2025.

\bibitem[Chen et~al.(2021)Chen, Tworek, Jun, Yuan, Pinto, Kaplan, Edwards,
  Burda, Joseph, Brockman, et~al.]{chen2021evaluating}
Mark Chen, Jerry Tworek, Heewoo Jun, Qiming Yuan, Henrique Ponde De~Oliveira
  Pinto, Jared Kaplan, Harri Edwards, Yuri Burda, Nicholas Joseph, Greg
  Brockman, et~al.
\newblock Evaluating large language models trained on code.
\newblock \emph{arXiv preprint arXiv:2107.03374}, 2021.

\bibitem[Chen et~al.(2025)Chen, Qin, Wu, Ling, Ye, Zhao, and
  Shi]{chen2025passk}
Zhipeng Chen, Xiaobo Qin, Youbin Wu, Yue Ling, Qinghao Ye, Wayne~Xin Zhao, and
  Guang Shi.
\newblock Pass@k training for adaptively balancing exploration and exploitation
  of large reasoning models.
\newblock \emph{arXiv preprint arXiv:2508.10751}, 2025.

\bibitem[Cheng et~al.(2026)Cheng, Huang, Zhu, Dai, Zhao, Zhang, and
  Wei]{cheng2026reasoning}
Daixuan Cheng, Shaohan Huang, Xuekai Zhu, Bo~Dai, Xin Zhao, Zhenliang Zhang,
  and Furu Wei.
\newblock Reasoning with exploration: An entropy perspective.
\newblock In \emph{Proceedings of the AAAI Conference on Artificial
  Intelligence}, volume~40, pages 30377--30385, 2026.

\bibitem[Fang et~al.(2026)Fang, Lin, Fu, Qin, Shi, Hu, Pan, Zeng, and
  Cai]{fang2026allocate}
Yangyi Fang, Jiaye Lin, Xiaoliang Fu, Cong Qin, Haolin Shi, Chaowen Hu, Lu~Pan,
  Ke~Zeng, and Xunliang Cai.
\newblock How to allocate, how to learn? dynamic rollout allocation and
  advantage modulation for policy optimization.
\newblock \emph{arXiv preprint arXiv:2602.19208}, 2026.

\bibitem[Gai et~al.(2025)Gai, Zeng, Zhang, and
  Raghunathan]{gai2025differential}
Jingchu Gai, Guanning Zeng, Huaqing Zhang, and Aditi Raghunathan.
\newblock Differential smoothing mitigates sharpening and improves {LLM}
  reasoning.
\newblock \emph{arXiv preprint arXiv:2511.19942}, 2025.

\bibitem[Guo et~al.(2025)Guo, Yang, Zhang, Song, Wang, Zhu, Xu, Zhang, Ma, Bi,
  et~al.]{guo2025deepseek}
Daya Guo, Dejian Yang, Haowei Zhang, Junxiao Song, Peiyi Wang, Qihao Zhu,
  Runxin Xu, Ruoyu Zhang, Shirong Ma, Xiao Bi, et~al.
\newblock {DeepSeek-R1} incentivizes reasoning in {LLMs} through reinforcement
  learning.
\newblock \emph{Nature}, 645\penalty0 (8081):\penalty0 633--638, 2025.

\bibitem[He et~al.(2025)He, Fried, and Welleck]{he2025rewarding}
Andre~Wang He, Daniel Fried, and Sean Welleck.
\newblock Rewarding the unlikely: Lifting {GRPO} beyond distribution
  sharpening.
\newblock In \emph{Proceedings of the 2025 Conference on Empirical Methods in
  Natural Language Processing}, pages 25559--25571, 2025.

\bibitem[Hendrycks et~al.(2021)Hendrycks, Burns, Kadavath, Arora, Basart, Tang,
  Song, and Steinhardt]{hendrycks2021measuring}
Dan Hendrycks, Collin Burns, Saurav Kadavath, Akul Arora, Steven Basart, Eric
  Tang, Dawn Song, and Jacob Steinhardt.
\newblock Measuring mathematical problem solving with the {MATH} dataset.
\newblock \emph{arXiv preprint arXiv:2103.03874}, 2021.

\bibitem[Liu et~al.(2025)Liu, Chen, Li, Qi, Pang, Du, Lee, and
  Lin]{liu2025understanding}
Zichen Liu, Changyu Chen, Wenjun Li, Penghui Qi, Tianyu Pang, Chao Du, Wee~Sun
  Lee, and Min Lin.
\newblock Understanding {R1-Zero}-like training: A critical perspective.
\newblock \emph{arXiv preprint arXiv:2503.20783}, 2025.

\bibitem[Nguyen et~al.(2026)Nguyen, Nguyen, Ma, Zhao, She, and
  Nguyen]{nguyen2026adaptive}
Hieu~Trung Nguyen, Bao Nguyen, Wenao Ma, Yuzhi Zhao, Ruifeng She, and Viet~Anh
  Nguyen.
\newblock Adaptive rollout allocation for online reinforcement learning with
  verifiable rewards.
\newblock \emph{arXiv preprint arXiv:2602.01601}, 2026.

\bibitem[Ospanov et~al.(2025)Ospanov, Feng, Sun, Bai, Shen, and
  Farnia]{ospanov2025hermes}
Azim Ospanov, Zijin Feng, Jiacheng Sun, Haoli Bai, Xin Shen, and Farzan Farnia.
\newblock {Hermes}: Towards efficient and verifiable mathematical reasoning in
  {LLMs}.
\newblock \emph{arXiv preprint arXiv:2511.18760}, 2025.

\bibitem[Patel et~al.(2024)Patel, Chakraborty, Suttle, Wang, Bedi, and
  Manocha]{patel2024aime}
Bhrij Patel, Souradip Chakraborty, Wesley~A Suttle, Mengdi Wang, Amrit~Singh
  Bedi, and Dinesh Manocha.
\newblock {AIME}: {AI} system optimization via multiple {LLM} evaluators.
\newblock \emph{arXiv preprint arXiv:2410.03131}, 2024.

\bibitem[Qu et~al.(2026{\natexlab{a}})Qu, Wang, Mao, Hu, Ommer, and
  Ji]{qu2026can}
Yun Qu, Qi~Wang, Yixiu Mao, Vincent~Tao Hu, Bj{\"o}rn Ommer, and Xiangyang Ji.
\newblock Can prompt difficulty be online predicted for accelerating {RL}
  finetuning of reasoning models?
\newblock In \emph{Proceedings of the 32nd ACM SIGKDD Conference on Knowledge
  Discovery and Data Mining V. 1}, pages 1240--1250, 2026{\natexlab{a}}.

\bibitem[Qu et~al.(2026{\natexlab{b}})Qu, Wang, Mao, Zou, Jiang, Liu, Bai,
  Yang, Chen, Yang, et~al.]{qu2026small}
Yun Qu, Qi~Wang, Yixiu Mao, Heming Zou, Yuhang Jiang, Weijie Liu, Clive Bai,
  Kai Yang, Yangkun Chen, Saiyong Yang, et~al.
\newblock Small generalizable prompt predictive models can steer efficient {RL}
  post-training of large reasoning models.
\newblock \emph{arXiv preprint arXiv:2602.01970}, 2026{\natexlab{b}}.

\bibitem[Schulman et~al.(2017)Schulman, Wolski, Dhariwal, Radford, and
  Klimov]{schulman2017proximal}
John Schulman, Filip Wolski, Prafulla Dhariwal, Alec Radford, and Oleg Klimov.
\newblock Proximal policy optimization algorithms.
\newblock \emph{arXiv preprint arXiv:1707.06347}, 2017.

\bibitem[Shao et~al.(2024)Shao, Wang, Zhu, Xu, Song, Bi, Zhang, Zhang, Li, Wu,
  et~al.]{shao2024deepseekmath}
Zhihong Shao, Peiyi Wang, Qihao Zhu, Runxin Xu, Junxiao Song, Xiao Bi, Haowei
  Zhang, Mingchuan Zhang, YK~Li, Yang Wu, et~al.
\newblock {DeepSeekMath}: Pushing the limits of mathematical reasoning in open
  language models.
\newblock \emph{arXiv preprint arXiv:2402.03300}, 2024.

\bibitem[Snell et~al.(2024)Snell, Lee, Xu, and Kumar]{snell2024scaling}
Charlie Snell, Jaehoon Lee, Kelvin Xu, and Aviral Kumar.
\newblock Scaling {LLM} test-time compute optimally can be more effective than
  scaling model parameters.
\newblock \emph{arXiv preprint arXiv:2408.03314}, 2024.

\bibitem[Song et~al.(2020)Song, Tan, Qin, Lu, and Liu]{song2020mpnet}
Kaitao Song, Xu~Tan, Tao Qin, Jianfeng Lu, and Tie-Yan Liu.
\newblock {MPNet}: Masked and permuted pre-training for language understanding.
\newblock \emph{Advances in Neural Information Processing Systems},
  33:\penalty0 16857--16867, 2020.

\bibitem[Tajwar et~al.(2026)Tajwar, Zeng, Zhou, Song, Arora, Jiang, Schneider,
  Salakhutdinov, Feng, and Zanette]{tajwar2026maximum}
Fahim Tajwar, Guanning Zeng, Yueer Zhou, Yuda Song, Daman Arora, Yiding Jiang,
  Jeff Schneider, Ruslan Salakhutdinov, Haiwen Feng, and Andrea Zanette.
\newblock Maximum likelihood reinforcement learning.
\newblock \emph{arXiv preprint arXiv:2602.02710}, 2026.

\bibitem[Wang et~al.(2025)Wang, Yu, Gao, Zheng, Liu, Lu, Dang, Chen, Yang,
  Zhang, et~al.]{wang2025beyond}
Shenzhi Wang, Le~Yu, Chang Gao, Chujie Zheng, Shixuan Liu, Rui Lu, Kai Dang,
  Xionghui Chen, Jianxin Yang, Zhenru Zhang, et~al.
\newblock Beyond the 80/20 rule: High-entropy minority tokens drive effective
  reinforcement learning for {LLM} reasoning.
\newblock \emph{arXiv preprint arXiv:2506.01939}, 2025.

\bibitem[Wen et~al.(2025)Wen, Liu, Zheng, Ye, Wu, Wang, Xu, Liang, Li, Miao,
  et~al.]{wen2025reinforcement}
Xumeng Wen, Zihan Liu, Shun Zheng, Shengyu Ye, Zhirong Wu, Yang Wang, Zhijian
  Xu, Xiao Liang, Junjie Li, Ziming Miao, et~al.
\newblock Reinforcement learning with verifiable rewards implicitly
  incentivizes correct reasoning in base {LLMs}.
\newblock \emph{arXiv preprint arXiv:2506.14245}, 2025.

\bibitem[Wu et~al.(2025)Wu, Xuan, Lu, Liu, Dong, Harchaoui, and
  Choi]{wu2025invisible}
Fang Wu, Weihao Xuan, Ximing Lu, Mingjie Liu, Yi~Dong, Zaid Harchaoui, and
  Yejin Choi.
\newblock The invisible leash: Why {RLVR} may or may not escape its origin.
\newblock \emph{arXiv preprint arXiv:2507.14843}, 2025.

\bibitem[Xu and Ding(2025)]{xu2025single}
Zhongwen Xu and Zihan Ding.
\newblock Single-stream policy optimization.
\newblock \emph{arXiv preprint arXiv:2509.13232}, 2025.

\bibitem[Yang et~al.(2025{\natexlab{a}})Yang, Yang, Zhang, Hui, Zheng, Yu, Li,
  Liu, Huang, Wei, et~al.]{yang2025qwen25}
An~Yang, Baosong Yang, Beichen Zhang, Binyuan Hui, Bo~Zheng, Bowen Yu,
  Chengyuan Li, Dayiheng Liu, Fei Huang, Haoran Wei, et~al.
\newblock {Qwen2.5} technical report.
\newblock \emph{arXiv preprint arXiv:2412.15115}, 2025{\natexlab{a}}.

\bibitem[Yang et~al.(2025{\natexlab{b}})Yang, Guo, Huang, Wang, Xie, Li, Wang,
  Liang, and Tang]{yang2025depth}
Zhicheng Yang, Zhijiang Guo, Yinya Huang, Yongxin Wang, Dongchun Xie, Hanhui
  Li, Yiwei Wang, Xiaodan Liang, and Jing Tang.
\newblock Depth-breadth synergy in {RLVR}: Unlocking {LLM} reasoning gains with
  adaptive exploration.
\newblock \emph{arXiv preprint arXiv:2508.13755}, 2025{\natexlab{b}}.

\bibitem[Yu et~al.(2025)Yu, Zhang, Zhu, Yuan, Zuo, Yue, Dai, Fan, Liu, Liu,
  et~al.]{yu2025dapo}
Qiying Yu, Zheng Zhang, Ruofei Zhu, Yufeng Yuan, Xiaochen Zuo, Yu~Yue, Weinan
  Dai, Tiantian Fan, Gaohong Liu, Lingjun Liu, et~al.
\newblock {DAPO}: An open-source {LLM} reinforcement learning system at scale.
\newblock \emph{arXiv preprint arXiv:2503.14476}, 2025.

\bibitem[Yue et~al.(2025)Yue, Chen, Lu, Zhao, Wang, Song, and
  Huang]{yue2025does}
Yang Yue, Zhiqi Chen, Rui Lu, Andrew Zhao, Zhaokai Wang, Shiji Song, and Gao
  Huang.
\newblock Does reinforcement learning really incentivize reasoning capacity in
  {LLMs} beyond the base model?
\newblock \emph{arXiv preprint arXiv:2504.13837}, 2025.

\bibitem[Zeng et~al.(2025)Zeng, Huang, Liu, Liu, He, Ma, and
  He]{zeng2025simplerl}
Weihao Zeng, Yuzhen Huang, Qian Liu, Wei Liu, Keqing He, Zejun Ma, and Junxian
  He.
\newblock {SimpleRL-Zoo}: Investigating and taming zero reinforcement learning
  for open base models in the wild.
\newblock \emph{arXiv preprint arXiv:2503.18892}, 2025.

\bibitem[Zhang et~al.(2025)Zhang, Yuan, Huang, You, Hu, Ruan, Chen, and
  Hu]{zhang2025revisiting}
Xiaoyun Zhang, Xiaojian Yuan, Di~Huang, Wang You, Chen Hu, Jingqing Ruan,
  Kejiang Chen, and Xing Hu.
\newblock Revisiting entropy regularization: Adaptive coefficient unlocks its
  potential for {LLM} reinforcement learning.
\newblock \emph{arXiv preprint arXiv:2510.10959}, 2025.

\bibitem[Zhu et~al.(2025)Zhu, Xia, Wei, Chen, Chen, and
  Meng]{zhu2025surprising}
Xinyu Zhu, Mengzhou Xia, Zhepei Wei, Wei-Lin Chen, Danqi Chen, and Yu~Meng.
\newblock The surprising effectiveness of negative reinforcement in {LLM}
  reasoning.
\newblock \emph{arXiv preprint arXiv:2506.01347}, 2025.

\end{thebibliography}

\newpage

\appendix

\section{Extended related work}
\label{app:related}

Compared with classical Proximal Policy Optimization (PPO)~\citep{schulman2017proximal}, group-based methods avoid learning a value function, which is expensive and unstable in large-scale LLM training, and they have been adopted in the training pipelines of powerful open-source LLMs~\citep{guo2025deepseek,yang2025qwen25}. Entropy collapse, commonly held responsible for Pass@$K$ degradation, refers to the policy becoming overly deterministic after RL optimization, which reduces output diversity and limits the benefit of multiple sampling~\citep{he2025rewarding,yue2025does}. Among the remedies discussed in Section~\ref{sec:related}, \cite{cheng2026reasoning} use the entropy term to encourage longer exploratory reasoning chains; \cite{wang2025beyond} apply RL updates only through the roughly 20\% of tokens with the highest entropy; and \cite{zhu2025surprising} upweight negative-sample reinforcement to suppress incorrect generations and redistribute probability mass toward alternative candidates.

\section{Proofs}
\subsection{Proof of Proposition~\ref{prop:greedy-optimal}}
\label{app:greedy-proof}

We first show that the marginal gains $M_i(0),M_i(1),\ldots$ for each prompt form a decreasing sequence.
For $P_i \mid C_i = c_i \sim \mathrm{Beta}(\alpha_i,\beta_i)$, the hit utility in~\eqref{eq:hit-utility} admits the closed form
\begin{align}
U_i^{\mathrm{hit}}(\Delta_i)
&=
1-\mathbb{E}_{P_i\mid C_i = c_i}\!\left[(1-P_i)^{\Delta_i}\right] \nonumber \\
&=
1-
\frac{1}{B(\alpha_i,\beta_i)}
\int_0^1 p^{\alpha_i-1}(1-p)^{\beta_i+\Delta_i-1}\,dp \nonumber \\
&=
1-\frac{B(\alpha_i,\beta_i+\Delta_i)}{B(\alpha_i,\beta_i)},
\label{eq:appendix-hit-closed-form}
\end{align}
where $B(a,b)=\int_0^1 p^{a-1}(1-p)^{b-1}\,dp$ denotes the beta function.

Recall that we define the marginal gain of the $(\ell+1)$-th additional rollout for prompt $i$ as
\begin{equation}
M_i(\ell)
:=
U_i^{\mathrm{hit}}(\ell+1)-U_i^{\mathrm{hit}}(\ell),
\qquad
\ell \in \mathbb{Z}_{\ge 0}.
\label{eq:appendix-marginal}
\end{equation}
Using~\eqref{eq:appendix-hit-closed-form}, we have
\begin{align}
M_i(\ell)
&=
\frac{B(\alpha_i,\beta_i+\ell)}{B(\alpha_i,\beta_i)}
-
\frac{B(\alpha_i,\beta_i+\ell+1)}{B(\alpha_i,\beta_i)} \nonumber \\
&=
\frac{B(\alpha_i+1,\beta_i+\ell)}{B(\alpha_i,\beta_i)},
\label{eq:appendix-marginal-closed-form}
\end{align}
where the last equality uses the fact that $B(a,b)=B(a+1,b)+B(a,b+1)$.

Note that the allocation objective can be written as a sum of marginal gains:
\begin{equation}
\sum_{i=1}^{\Gamma} U_i^{\mathrm{hit}}(\Delta_i)
=
\sum_{i=1}^{\Gamma} \sum_{\ell=0}^{\Delta_i-1} M_i(\ell).
\label{eq:appendix-objective}
\end{equation}
Moreover, each marginal sequence is strictly decreasing:
\begin{equation}
\frac{M_i(\ell+1)}{M_i(\ell)}
=
\frac{B(\alpha_i+1,\beta_i+\ell+1)}
     {B(\alpha_i+1,\beta_i+\ell)}
=
\frac{\beta_i+\ell}{\alpha_i+\beta_i+\ell+1}
< 1,
\label{eq:appendix-ratio}
\end{equation}
since $\alpha_i>0$. Hence
\begin{equation}
M_i(0) > M_i(1) > M_i(2) > \cdots > 0.
\label{eq:appendix-monotone}
\end{equation}

We next prove the optimality of the greedy allocation rule by comparing its objective value with that of any feasible allocation. Let $\Delta^{\mathrm{g}}=(\Delta^{\mathrm{g}}_1,\ldots,\Delta^{\mathrm{g}}_{\Gamma})$ be the allocation returned by the greedy procedure, which repeatedly assigns the next rollout to the prompt with the largest current marginal gain. By the greedy property, we have the following condition:
\begin{equation}
M_i(\Delta_i^{\mathrm{g}}-1)
\ge
M_j(\Delta_j^{\mathrm{g}})
\qquad
\text{for every } i \text{ with } \Delta_i^{\mathrm{g}}>0 \text{ and every } j.
\label{eq:appendix-exchange-condition}
\end{equation}
Indeed, when greedy assigned the last rollout to prompt $i$, the available marginal for prompt $i$ was $M_i(\Delta_i^{\mathrm{g}}-1)$. If prompt $j$ had already received $r_j$ rollouts at that time, then $r_j \le \Delta_j^{\mathrm{g}}$. By monotonicity, $M_j(r_j)\ge M_j(\Delta_j^{\mathrm{g}})$, and since greedy selected prompt $i$ at that step,
\[
M_i(\Delta_i^{\mathrm{g}}-1)
\ge
M_j(r_j)
\ge
M_j(\Delta_j^{\mathrm{g}}).
\]

Now consider any feasible allocation $\Delta=(\Delta_1,\ldots,\Delta_{\Gamma})$ satisfying
\[
\sum_{i=1}^{\Gamma} \Delta_i = \Gamma(G-G_0).
\]
Define
\[
A=\{i:\Delta_i>\Delta_i^{\mathrm{g}}\},
\qquad
D=\{j:\Delta_j<\Delta_j^{\mathrm{g}}\}.
\]
Because $\Delta$ and $\Delta^{\mathrm{g}}$ use the same total budget, we have
\[
\sum_{i\in A}(\Delta_i-\Delta_i^{\mathrm{g}})
=
\sum_{j\in D}(\Delta_j^{\mathrm{g}}-\Delta_j).
\]
Relative to the greedy allocation, prompts in $A$ receive extra marginal gains
\[
M_i(\Delta_i^{\mathrm{g}}),\,
M_i(\Delta_i^{\mathrm{g}}+1),\,
\ldots,\,
M_i(\Delta_i-1),
\]
each of which is at most $M_i(\Delta_i^{\mathrm{g}})$. Prompts in $D$ lose marginal gains
\[
M_j(\Delta_j),\,
M_j(\Delta_j+1),\,
\ldots,\,
M_j(\Delta_j^{\mathrm{g}}-1),
\]
each of which is at least $M_j(\Delta_j^{\mathrm{g}}-1)$.

By the condition in~\eqref{eq:appendix-exchange-condition}, every extra marginal gain added by $\Delta$ is no larger than some marginal gain removed from $\Delta^{\mathrm{g}}$. Therefore,
\[
\sum_{i=1}^{\Gamma} \sum_{\ell=0}^{\Delta_i-1} M_i(\ell)
\le
\sum_{i=1}^{\Gamma} \sum_{\ell=0}^{\Delta_i^{\mathrm{g}}-1} M_i(\ell).
\]
Using~\eqref{eq:appendix-objective}, this implies
\[
\sum_{i=1}^{\Gamma} U_i^{\mathrm{hit}}(\Delta_i)
\le
\sum_{i=1}^{\Gamma} U_i^{\mathrm{hit}}(\Delta_i^{\mathrm{g}}).
\]
Thus no feasible allocation has larger objective value than the greedy allocation, so the greedy allocator exactly solves~\eqref{eq:allocation-problem}.

\subsection{Degeneracy of Plug-in Hit Utility}
\label{app:plugin-degeneracy}
Here, we explain the reason that we use the posterior distribution to model the success probability $p_i$
rather than the plug-in estimate $\hat p_i=C_i/G_0$. Using the point estimate, the corresponding hit utility will be defined as
\[
\widetilde U_i^{\mathrm{hit}}(\Delta_i)=1-(1-\hat p_i)^{\Delta_i}.
\]
The key difference is that our Bayesian formulation~\eqref{eq:hit-utility} computes the expectation over the posterior $P_i |C_i =  c_i$. If $c_i=0$, then $\hat p_i=0$ and hence $\widetilde U_i^{\mathrm{hit}}(\Delta_i)=0$ for all $\Delta_i\ge 0$. Its marginal gain is therefore
\[
\widetilde M_i(\ell)
=
\widetilde U_i^{\mathrm{hit}}(\ell+1)
-
\widetilde U_i^{\mathrm{hit}}(\ell)
=0,
\qquad \forall \ell\ge 0.
\]
For any prompt $j$ with $0<\hat p_j<1$, the marginal gain is
\[
\widetilde M_j(\ell)=\hat p_j(1-\hat p_j)^\ell>0,
\qquad \forall \ell\ge 0.
\]
Thus, under the greedy allocation rule, a prompt with $c_i=0$ is never selected while any prompt with positive marginal gain remains. Intuitively, the plug-in estimator treats a zero-count prompt as if its success probability were exactly zero. The greedy allocation rule therefore withholds additional rollouts from such prompts, which conflicts with the goal of continuing to explore challenging prompts whose true success probability may be small but nonzero.

\section{HORA pseudocode}
\label{app:algorithm}

Algorithm~\ref{alg:hora} summarizes the rollout construction of HORA for one training step (Section~\ref{sec:HORA}).

\begin{algorithm}[ht]
\caption{HORA rollout construction for one training step.}
\label{alg:hora}
\begin{algorithmic}[1]
\Require batch of prompts $\{q_i\}_{i=1}^\Gamma$; group size $G$; pre-rollout count $G_0 < G$; prior $(\alpha_0, \beta_0)$; behavior policy $\pi_{\theta_{\mathrm{old}}}$
\Ensure variable-size rollout groups $\{(o_{i,1}, \ldots, o_{i,G_i})\}_{i=1}^\Gamma$
\Statex \textbf{Phase~A: pre-rollouts and posterior}
\For{$i = 1, \ldots, \Gamma$}
  \State draw $G_0$ pre-rollouts $\{o_{i,j}^{\mathrm{pre}}\}_{j=1}^{G_0} \sim \pi_{\theta_{\mathrm{old}}}(\cdot \mid q_i)$
  \State $c_i \gets \sum_{j=1}^{G_0} R_{\mathrm{acc}}(q_i, o_{i,j}^{\mathrm{pre}})$
  \State $(\alpha_i, \beta_i) \gets (\alpha_0 + c_i,\; \beta_0 + G_0 - c_i)$
  \State $\Delta_i \gets 0$
\EndFor
\Statex \textbf{Phase~B: greedy hit-utility allocation}
\State $\mathcal{B} \gets \Gamma(G - G_0)$
\For{$b = 1, \ldots, \mathcal{B}$}
  \State $i^\star \gets \arg\max_{i} \; B(\alpha_i + 1,\, \beta_i + \Delta_i) / B(\alpha_i, \beta_i)$
  \State $\Delta_{i^\star} \gets \Delta_{i^\star} + 1$
\EndFor
\Statex \textbf{Additional rollouts and group assembly}
\For{$i = 1, \ldots, \Gamma$}
  \State draw $\Delta_i$ additional rollouts $\{o_{i,j}^{\mathrm{add}}\}_{j=1}^{\Delta_i} \sim \pi_{\theta_{\mathrm{old}}}(\cdot \mid q_i)$
  \State pool $\{o_{i,j}^{\mathrm{pre}}\}_{j=1}^{G_0}$ and $\{o_{i,j}^{\mathrm{add}}\}_{j=1}^{\Delta_i}$ into $\{o_{i,j}\}_{j=1}^{G_i}$, with $G_i = G_0 + \Delta_i$
\EndFor
\State \Return $\{(o_{i,1}, \ldots, o_{i,G_i})\}_{i=1}^\Gamma$
\end{algorithmic}
\end{algorithm}

\section{Training configurations}
\label{app:configs}

Tables~\ref{tab:config-1.5B}--\ref{tab:config-7B} list the per-model HORA
training configurations used for the main-results runs in
Table~\ref{tab:main-results}. GRPO baselines use identical hyperparameters
except for the HORA-specific rows (pre-rollout count $G_0$ and Beta prior).
All runs use the TRL implementation of GRPO with vLLM rollout in colocate
mode. To accelerate format learning, we run vanilla GRPO as a warm-up at
the start of each HORA run and switch to HORA once the global
format-reward mean stays at or above $0.85$ for three consecutive
generation cycles; in practice all three models cross this threshold within
about $10$ optimizer steps ($\lesssim 5\%$ of total training). Phase~B
allocation is run on four mini-batches of $\Gamma /4$ prompts rather than the
full batch in one shot. Thus the optimality result in Proposition~\ref{prop:greedy-optimal} applies within each allocation shard, while the total rollout budget over the full training batch remains unchanged. Rows that match TRL defaults (linear LR scheduler, max grad norm $1.0$, weight decay $0$, entropy coefficient
$0$) are omitted.

\begin{table}[ht]
\centering
\caption{Configuration for Qwen2.5-1.5B-Instruct.}
\label{tab:config-1.5B}
\small
\begin{tabular}{llll}
\toprule
\textbf{Parameter} & \textbf{Value} & \textbf{Parameter} & \textbf{Value} \\
\midrule
Pretrained model            & Qwen2.5-1.5B-Instruct & Training set                    & MATH12k                \\
Prompts per batch $\Gamma$       & 60                    & Generations per prompt $G$      & 32                     \\
Pre-rollout count $G_0$     & 8                     & Beta prior $(\alpha_0,\beta_0)$ & $(1,1)$                \\
Training steps              & 200                   & Learning rate                   & $1.0 \times 10^{-6}$   \\
Clip ratio $\varepsilon$    & 0.2                   & KL coefficient           & 0.0                    \\
Loss type                   & Dr.GRPO               & Importance sampling             & token-truncate         \\
Max prompt length           & 2048                  & Max response length             & 2048                   \\
Reward weights (fmt / acc)  & 0.5 / 0.5             & Gradient updates per RL step    & 1                      \\
Rollout temperature         & 0.7                   & Precision                       & bfloat16               \\
Rollout engine              & vLLM (colocate)       & Device                          & 3 $\times$ Nvidia-A100 \\
\bottomrule
\end{tabular}
\end{table}

\begin{table}[ht]
\centering
\caption{Configuration for Qwen2.5-3B.}
\label{tab:config-3B}
\small
\begin{tabular}{llll}
\toprule
\textbf{Parameter} & \textbf{Value} & \textbf{Parameter} & \textbf{Value} \\
\midrule
Pretrained model            & Qwen2.5-3B            & Training set                    & MATH12k                \\
Prompts per batch $\Gamma$       & 60                    & Generations per prompt $G$      & 32                     \\
Pre-rollout count $G_0$     & 8                     & Beta prior $(\alpha_0,\beta_0)$ & $(1,1)$                \\
Training steps              & 200                   & Learning rate                   & $1.0 \times 10^{-6}$   \\
Clip ratio $\varepsilon$    & 0.2                   & KL coefficient           & 0.0                    \\
Loss type                   & Dr.GRPO               & Importance sampling             & token-truncate         \\
Max prompt length           & 2048                  & Max response length             & 2048                   \\
Reward weights (fmt / acc)  & 0.5 / 0.5             & Gradient updates per RL step    & 1                      \\
Rollout temperature         & 0.7                   & Precision                       & bfloat16               \\
Rollout engine              & vLLM (colocate)       & Device                          & 3 $\times$ Nvidia-A100 \\
\bottomrule
\end{tabular}
\end{table}

\begin{table}[ht]
\centering
\caption{Configuration for Qwen2.5-7B.}
\label{tab:config-7B}
\small
\begin{tabular}{llll}
\toprule
\textbf{Parameter} & \textbf{Value} & \textbf{Parameter} & \textbf{Value} \\
\midrule
Pretrained model            & Qwen2.5-7B            & Training set                    & MATH12k                \\
Prompts per batch $\Gamma$       & 60                    & Generations per prompt $G$      & 32                     \\
Pre-rollout count $G_0$     & 8                     & Beta prior $(\alpha_0,\beta_0)$ & $(1,1)$                \\
Training steps              & 200                   & Learning rate                   & $1.0 \times 10^{-6}$   \\
Clip ratio $\varepsilon$    & 0.2                   & KL coefficient           & 0.0                    \\
Loss type                   & Dr.GRPO               & Importance sampling             & token-truncate         \\
Max prompt length           & 2048                  & Max response length             & 2048                   \\
Reward weights (fmt / acc)  & 0.5 / 0.5             & Gradient updates per RL step    & 1                      \\
Rollout temperature         & 0.7                   & Precision                       & bfloat16               \\
Rollout engine              & vLLM (colocate)       & Device                          & 6 $\times$ Nvidia-A100 \\
\bottomrule
\end{tabular}
\end{table}

\paragraph{Llama-3.2 and Qwen3 backbones.}
Table~\ref{tab:config-other} lists the training settings for Llama-3.2-3B-Instruct and Qwen3-1.7B-Base. Hyperparameters not listed there are the same as in Tables~\ref{tab:config-1.5B}--\ref{tab:config-7B}, and GRPO and HORA are trained with the same settings. Qwen3-1.7B-Base does not produce the \texttt{<think>} token, which is a special token in its vocabulary, so we prompt it to write its reasoning inside \texttt{<reasoning>} tags, and the format reward checks these tags instead. For Llama-3.2-3B-Instruct, we follow VIP~\citep{nguyen2026adaptive}, which trains the same model, and use only the accuracy reward, with the final answer given in \texttt{\textbackslash boxed\{\}}. Without a format reward, the format-based warm-up rule above does not apply, so we run the GRPO warm-up for a fixed $10$ optimizer steps.

\begin{table}[ht]
\centering
\caption{Training settings of the Llama-3.2 and Qwen3 backbones.}
\label{tab:config-other}
\small
\begin{tabular}{lccc}
\toprule
 & Llama-3.2-3B-Inst. & Qwen3-1.7B-Base & Qwen3-1.7B-Base \\
\midrule
Training set & MATH12k & MATH12k & DAPO-17k (12k subset) \\
Reasoning format & \texttt{<think>} + \texttt{\textbackslash boxed\{\}} & \texttt{<reasoning>}/\texttt{<answer>} & \texttt{<reasoning>}/\texttt{<answer>} \\
Reward (fmt / acc) & 0 / 1 & 0.5 / 0.5 & 0.5 / 0.5 \\
Max response length & 2048 & 2048 & 3072 \\
Learning rate & $1.0\times10^{-6}$ & $1.0\times10^{-6}$ & $1.5\times10^{-6}$ \\
Device & 6 $\times$ Nvidia-A100 & 3 $\times$ Nvidia-A100 & 2 $\times$ Nvidia-B200 \\
\bottomrule
\end{tabular}
\end{table}

\section{Additional Pass@K curves}
\label{app:full-passk}

\begin{figure}[!ht]
\centering
\includegraphics[width=\linewidth]{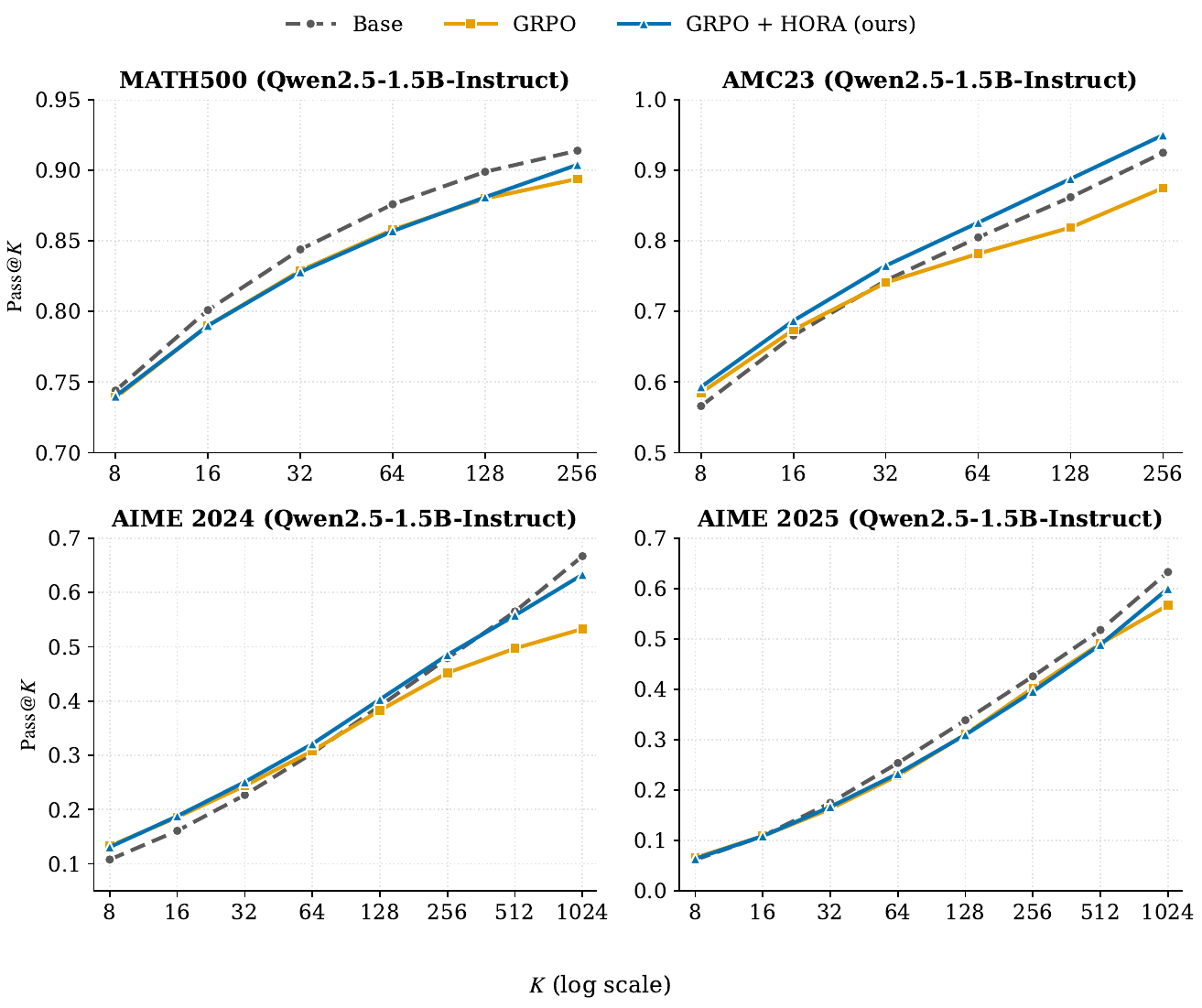}
\caption{Pass@$K$ curves for Qwen2.5-1.5B-Instruct across all four benchmarks.}
\label{fig:full-passk-1.5B}
\end{figure}

\begin{figure}[!ht]
\centering
\includegraphics[width=\linewidth]{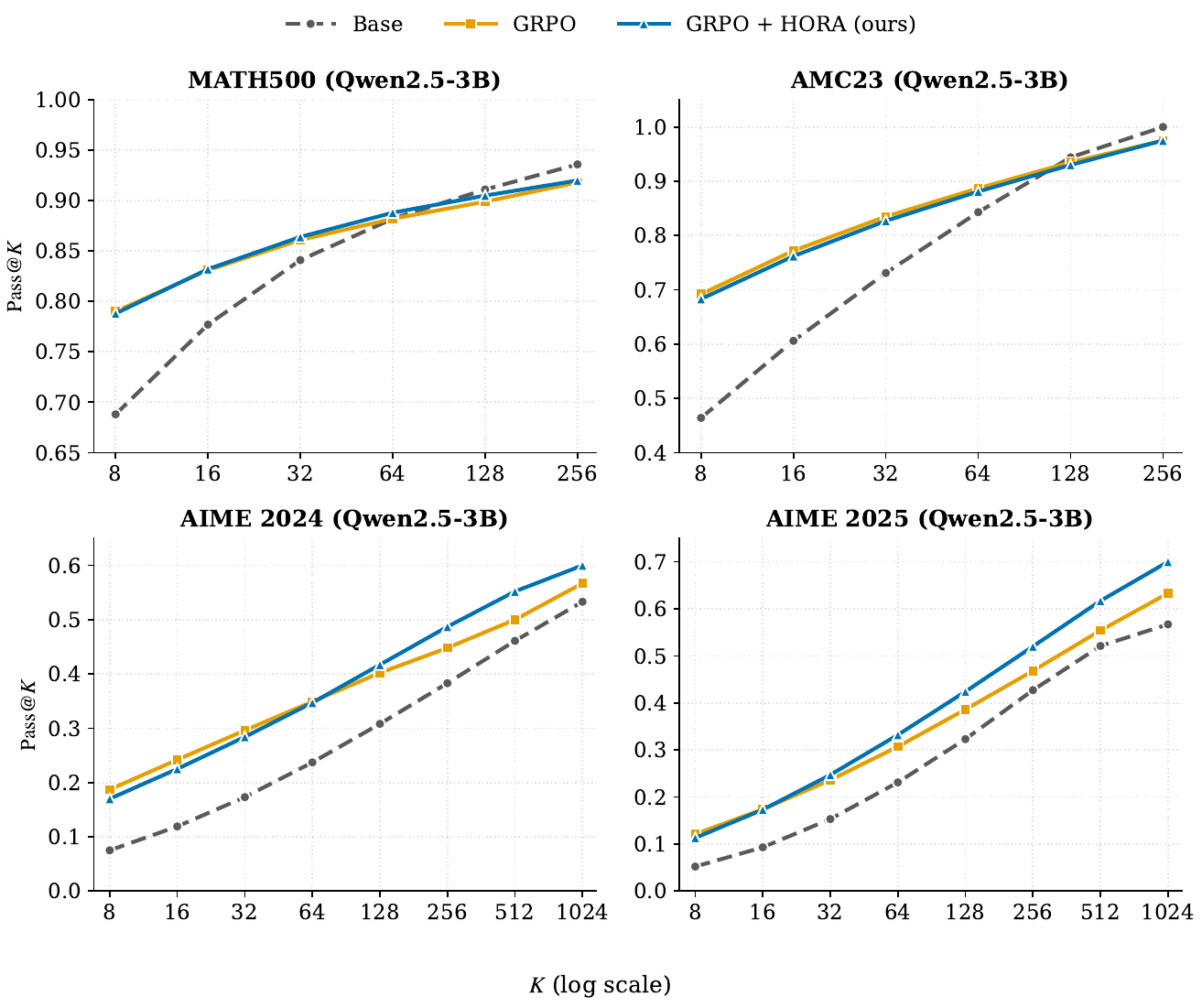}
\caption{Pass@$K$ curves for Qwen2.5-3B across all four benchmarks.}
\label{fig:full-passk-3B}
\end{figure}

\begin{figure}[!ht]
\centering
\includegraphics[width=\linewidth]{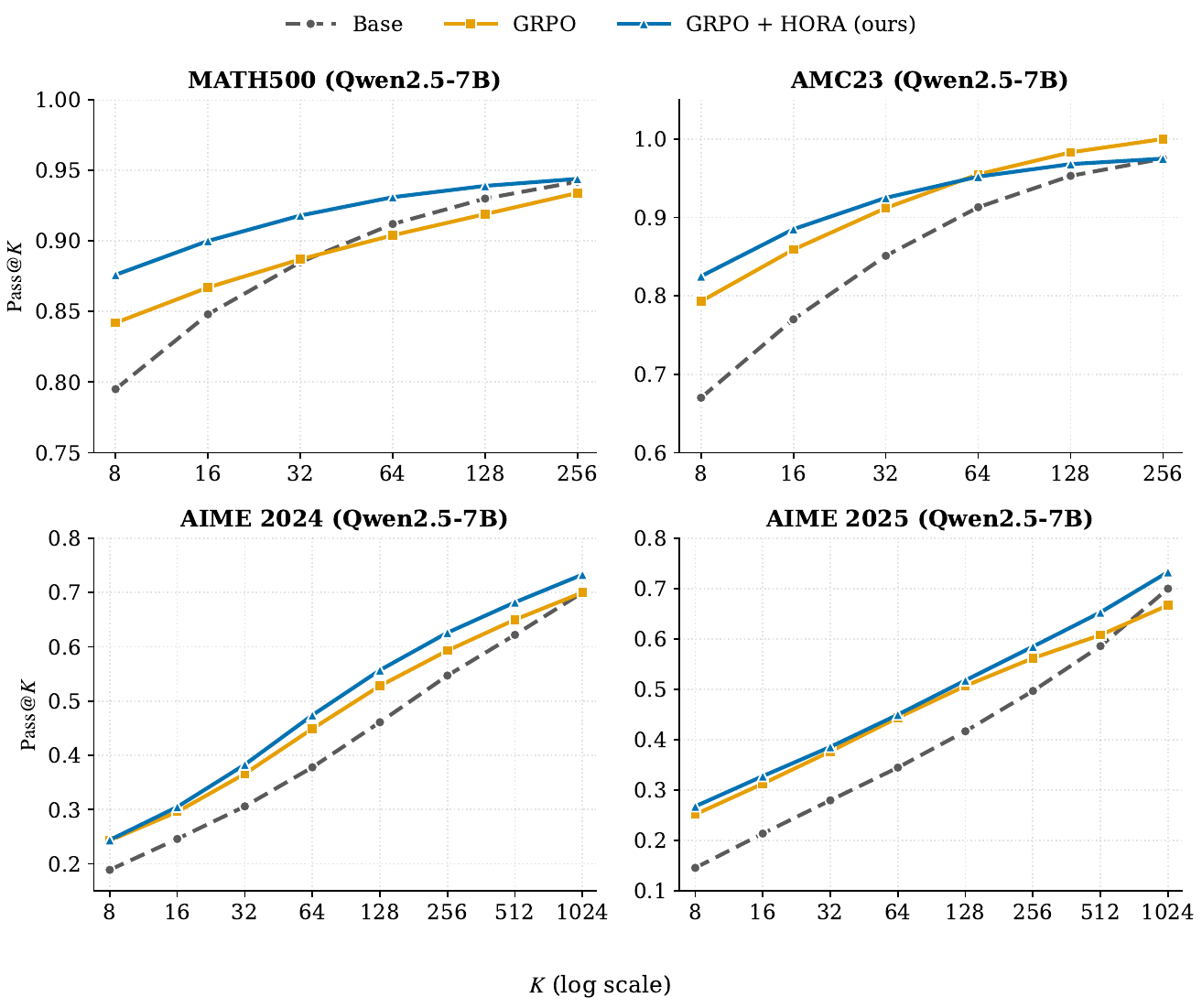}
\caption{Pass@$K$ curves for Qwen2.5-7B across all four benchmarks. The MATH500
and AIME~2025 panels reproduce the top row of Figure~\ref{fig:passk-aime}.}
\label{fig:full-passk-7B}
\end{figure}

\begin{figure}[!ht]
\centering
\includegraphics[width=\linewidth]{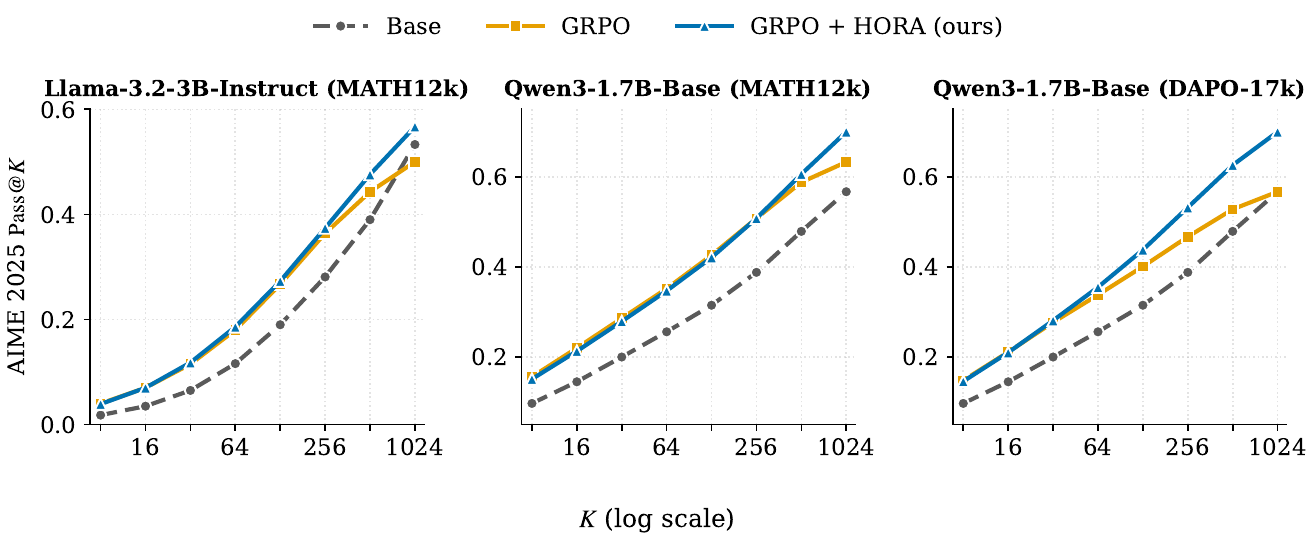}
\caption{AIME~2025 Pass@$K$ curves for Llama-3.2-3B-Instruct and Qwen3-1.7B-Base trained on MATH12k, and Qwen3-1.7B-Base trained on DAPO-17k.}
\label{fig:full-passk-families}
\end{figure}

Figures~\ref{fig:full-passk-1.5B}--\ref{fig:full-passk-families} report the full Pass@$K$ curves for the settings in Tables~\ref{tab:main-results} and~\ref{tab:families}.

\section{Length and entropy statistics}
\label{app:length-stats}

\begin{figure}[ht]
\centering
\includegraphics[width=\linewidth]{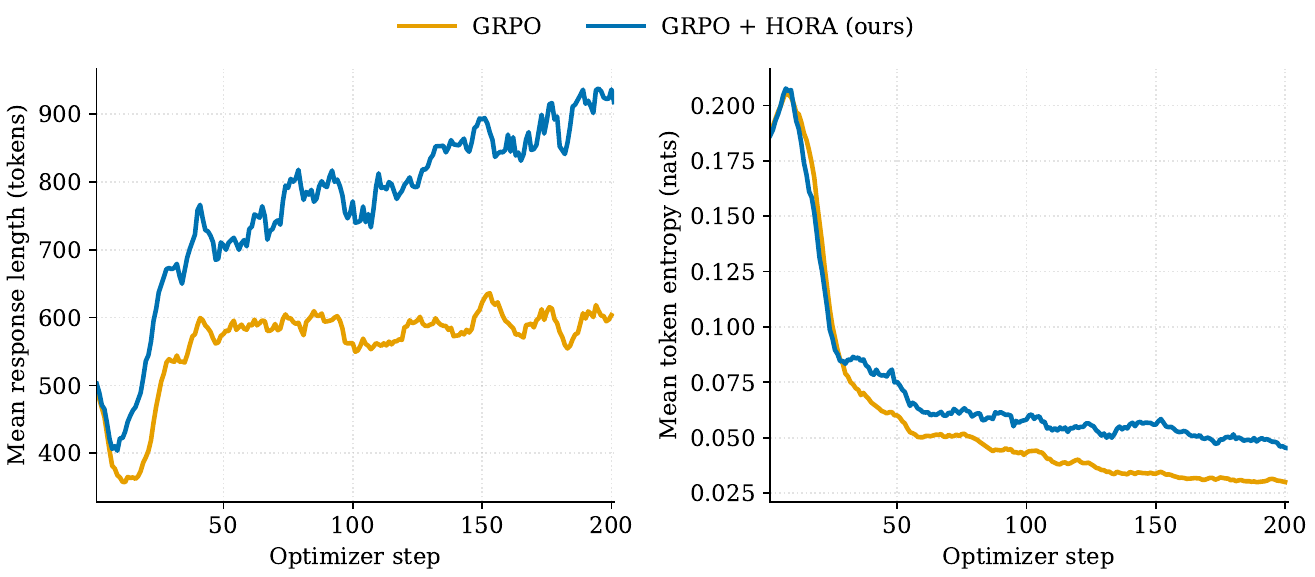}
\caption{Mean response length (left) and mean token entropy (right) over
training for GRPO + HORA and the GRPO baseline on Qwen2.5-7B. Both runs use
the same total rollout budget. Curves are smoothed with a centered moving
average of window $\max(5, n/30)$ where $n$ is the number of optimizer
steps logged; the smoother shrinks its window at the boundaries to avoid
edge artifacts.}
\label{fig:length-entropy}
\end{figure}

To corroborate the claim in Section~\ref{sec:main-results} that HORA
preserves more exploratory sampling behavior, Figure~\ref{fig:length-entropy}
plots mean response length and mean token entropy over training for the
Qwen2.5-7B HORA run against the
compute-matched GRPO baseline. HORA produces
noticeably longer completions throughout training ($\approx 300$-token
gap) and maintains a heavier entropy tail after the initial format-collapse
transient, indicating that the policy continues to allocate non-trivial
probability mass to alternative continuations.

\section{Prior-design predictor details}
\label{app:prior-details}
\begin{figure}[!ht]
\centering
\includegraphics[width=\linewidth]{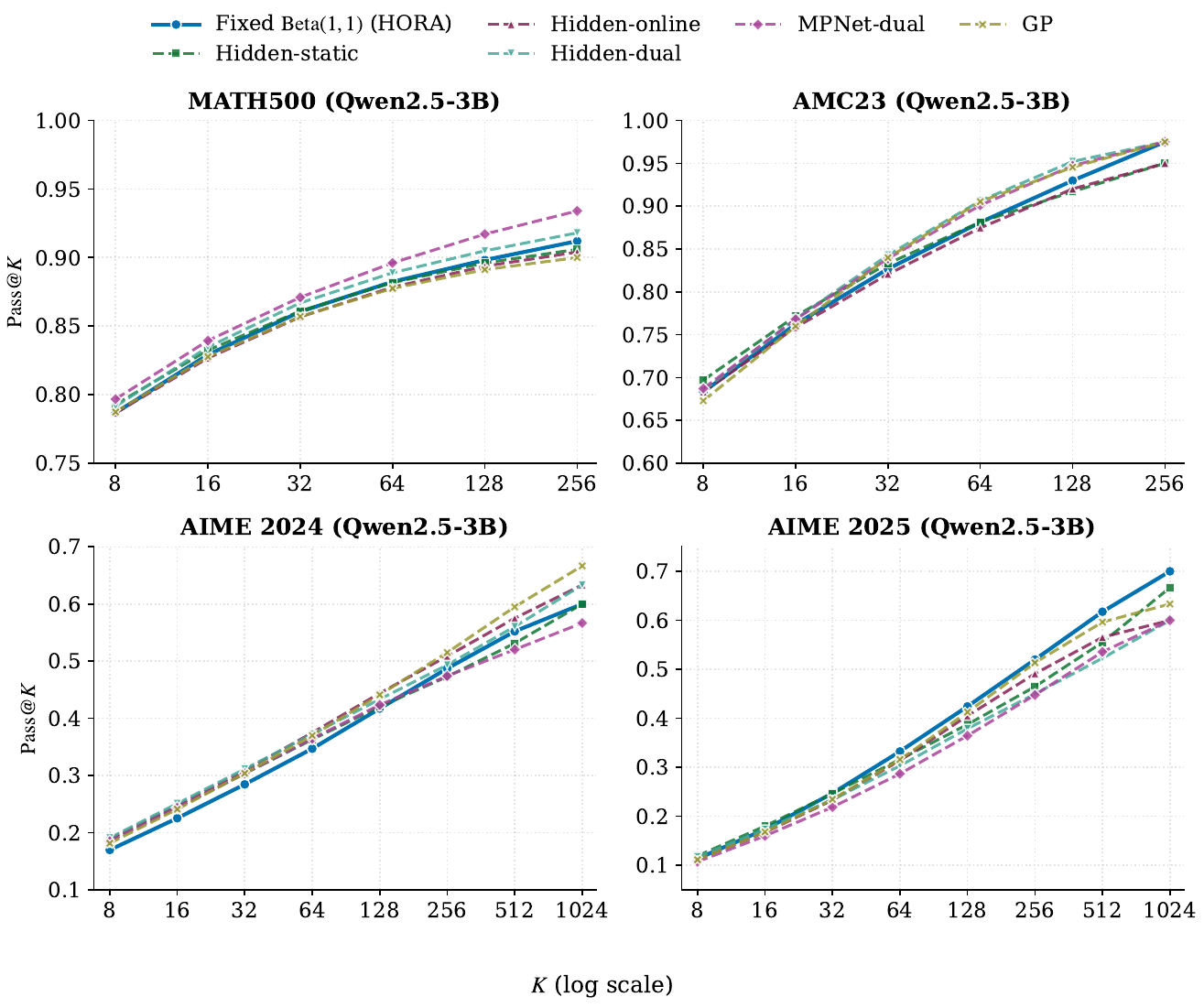}
\caption{Prior ablation across all four benchmarks on Qwen2.5-3B. The fixed
$\mathrm{Beta}(1,1)$ prior used by HORA (solid blue) is compared against the
five learned-prior variants described below. No learned variant uniformly
dominates the fixed prior across all four benchmarks: on AIME~2025 the
fixed prior is the strongest at $K = 1024$; on MATH500 the MPNet-dual
variant pulls slightly ahead at large $K$; individual variants take the
lead on different intermediate $K$ values, but no consistent ranking
emerges across benchmarks. This supports the negative finding in
Section~\ref{sec:ablations}: the same-step pre-rollout signal already
saturates the information available for allocation in this regime.}
\label{fig:prior-full}
\end{figure}

This section specifies the five learned-prior variants used in the prior
ablation of Section~\ref{sec:ablations}. All variants replace the fixed
$\mathrm{Beta}(1,1)$ prior used by HORA with a prompt-conditioned predictor
that produces a per-prompt success-rate estimate $\hat p_i$ (and, for the
dual-head variants, a per-prompt confidence $s_i$). The Beta posterior in~\eqref{eq:posterior} is then formed as
\begin{equation}
\alpha_i = s_i\, \hat p_i + c_i,
\qquad
\beta_i  = s_i\, (1 - \hat p_i) + G_0 - c_i,
\label{eq:posterior-learned}
\end{equation}
which collapses to the HORA fixed prior when $s_i = 2$ and $\hat p_i = 0.5$.
Single-head variants and the Gaussian Process (GP) \citep{nguyen2026adaptive} use a fixed $s_i = 2$, while the two
dual-head variants learn $s_i$ jointly with $\hat p_i$. All variants are
trained on the same Qwen2.5-3B / MATH12k pipeline as the main HORA run; only
the prior is changed. Figure~\ref{fig:prior-full} reports Pass@$K$ across
all four benchmarks.

\paragraph{Hidden-state linear probe (static / online / dual).}
The hidden-state variants extract the last-token hidden state at the
$\lfloor 0.6 L \rfloor$-th transformer block (mid-depth, $L =$ number of
hidden layers) of the frozen base model backbone, pass it through a LayerNorm, and
predict $\hat p_i$ with a linear head. The probe is pretrained offline on
held-out rollouts using the per-prompt binomial NLL
\[
\mathcal{L}_{\mathrm{bin}}
= - c_i \log \hat p_i - (G_0 - c_i) \log(1 - \hat p_i).
\]
Three update modes are evaluated:
\textbf{Hidden-static} keeps the pretrained probe frozen during RL training
(no online updates). \textbf{Hidden-online} updates the probe online: each
optimizer step replays the most recent $10$ generation cycles, takes $5$ Adam
steps with learning rate $10^{-3}$, weight decay $10^{-2}$, geometric
importance weight $\gamma^t = 0.97^t$, and maintains an EMA shadow with decay
$0.99$ that is used at inference. \textbf{Hidden-dual} replaces the
single-head probe with a dual-head probe that outputs $(\hat p_i, s_i)$ via
two parallel linear heads; $s_i$ is parametrized through a softplus and
clamped to $[0.25, 16]$. The probe is trained with the per-prompt
Beta-Binomial NLL
\[
\mathcal{L}_{\mathrm{BB}}
= -\log
\frac{B(c_i + s_i \hat p_i,\, G_0 - c_i + s_i(1 - \hat p_i))}
     {B(s_i \hat p_i,\, s_i (1 - \hat p_i))},
\]
using the same online schedule as Hidden-online.

\paragraph{MPNet-dual.}
Replaces the base policy hidden state with frozen sentence embeddings from
\texttt{all-\allowbreak mpnet-\allowbreak base-\allowbreak v2}~\citep{song2020mpnet}. The probe architecture
(LayerNorm + dual linear heads), Beta-Binomial loss, and online update
schedule are identical to Hidden-dual; only the input feature is different.

\paragraph{GP.}
We adopt the recursive Gaussian-process predictor proposed by
VIP~\citep{nguyen2026adaptive}: an RBF kernel over MPNet features with
median-distance bandwidth, prior mean $\mu_0 = -1$ in logit-space (so the
prior $\hat p_i \approx 0.27$), and recursive Bayesian updates on the
observed empirical rates $c_i / G_0$. We use the implementation provided by
the original authors and refer the reader to that paper \citep{nguyen2026adaptive} for the full
formulation.

\section{Compatibility with the RLOO advantage estimator}
\label{app:rloo-compatibility}

Table~\ref{tab:modularity} reports the per-benchmark Pass@1 and Pass@$K$ results of RLOO \citep{ahmadian2024back} and RLOO + HORA discussed in
Section~\ref{sec:main-results}. Both methods use the same total rollout
budget with $G=32$ on Qwen2.5-7B, and differ only in the within-group advantage estimator: the RLOO baseline applies the leave-one-out advantage \citep{ahmadian2024back} to a fixed-group rollout of size $G=32$, while
RLOO + HORA replaces the rollout stage with HORA's two-phase
posterior-guided allocator and uses the same RLOO advantage calculation. The
pattern observed in Section~\ref{sec:main-results} recurs here: HORA improves Pass@$K$ on three of four benchmarks while matching it
on the fourth, and matches the RLOO baseline on average Pass@1.

\begin{table}[ht]
\centering
\caption{Pass@1 and Pass@$K$ (\%) for RLOO and RLOO + HORA on Qwen2.5-7B.
Both methods use the same total rollout budget with $G=32$. We use $N=256$
for MATH500 and AMC23, and $N=1024$ for AIME~2024 and AIME~2025. Bold marks
the better value in each column.}
\label{tab:modularity}
\vspace{4pt}
\small
\setlength{\tabcolsep}{6pt}
\begin{tabular}{lcccccccc}
\toprule
\multirow{2}{*}{Method}
 & \multicolumn{2}{c}{MATH500} & \multicolumn{2}{c}{AMC23} & \multicolumn{2}{c}{AIME~2024} & \multicolumn{2}{c}{AIME~2025} \\
\cmidrule(lr){2-3} \cmidrule(lr){4-5} \cmidrule(lr){6-7} \cmidrule(lr){8-9}
 & Pass@1 & Pass@256 & Pass@1 & Pass@256 & Pass@1 & Pass@1024 & Pass@1 & Pass@1024 \\
\midrule
RLOO baseline    & 71.0          & 92.0          & 49.6          & 97.5           & \textbf{12.3} & 70.0          & \textbf{10.2} & \textbf{73.3} \\
RLOO + HORA      & \textbf{71.9} & \textbf{94.2} & \textbf{51.1} & \textbf{100.0} & 11.6          & \textbf{73.3} & 8.8           & \textbf{73.3} \\
\bottomrule
\end{tabular}
\end{table}

\section{Computational cost}
\label{app:cost}
Section~\ref{sec:main-results} reports the end-to-end wall-clock cost of HORA; here we break it down by stage. Each pair of GRPO and HORA runs uses the same hardware and batch configuration. At 3B, one generation cycle takes $19.2$\,s for GRPO, which generates all rollouts in a single stage, and $12.5$\,s plus $18.7$\,s for the two stages of HORA. Because the forward and backward passes take a large part of each optimizer step, the end-to-end difference is smaller: $427$\,s versus $457$\,s per step ($+6.9\%$), or $23.8$ versus $25.3$ hours for $200$ steps. At 7B, the two methods take $591$\,s and $582$\,s per step, so we see no measurable overhead. The greedy allocator takes $21$\,ms per step.

Two factors keep the extra cost small. First, vLLM uses continuous batching, so sequences that finish early free their slots, and each stage only waits for its longest completion. This wait is bounded by the $2048$-token response cap, which only $1\%$ to $5\%$ of completions reach in the main runs. Second, our HORA implementation drops groups with zero reward variance before the forward and backward passes, since they give no gradient ($18$ of $60$ groups per step on average at 3B), whereas the GRPO implementation still processes them. With much longer responses, the extra synchronization would take a larger share of the step.

\end{document}